\documentclass[letterpaper, 10 pt, conference]{ieeeconf}  

\IEEEoverridecommandlockouts                    

\usepackage[left=1.88cm,right=1.88cm,top=1.88cm,bottom=1.88cm]{geometry}
\usepackage{amsmath,amssymb,amsfonts}
\usepackage{cite}
\usepackage{graphicx}
\usepackage{textcomp}
\usepackage{xcolor}
\usepackage{dsfont}
\let\labelindent\relax
\usepackage{enumitem}
\usepackage{tabularx}
\usepackage{multirow}
\usepackage[dvipsnames]{xcolor}
\usepackage{times}
\usepackage{multicol}
\usepackage{booktabs} 

\usepackage{graphicx}
\usepackage{bm}
\usepackage[nolist]{acronym}

\usepackage[linesnumbered,ruled]{algorithm2e}

\usepackage{mathtools}
\usepackage{array}
\usepackage{dblfloatfix}
\usepackage{subcaption}
\usepackage{diagbox}
\usepackage{etoolbox}\AtBeginEnvironment{algorithmic}{\small}

\usepackage{color,soul} 

\newif\ifshowcomments
\showcommentstrue  

\ifshowcomments
    \newcommand{\bae}[1]{\hl{[SB: #1]}\protect\color{black}} 
    \newcommand{\di}[1]{\hl{[DI: #1]}\protect\color{black}} 
    \newcommand{\ft}[1]{\hl{[FT: #1]}\protect\color{black}} 
    \newcommand{\jd}[1]{\hl{[JD: #1]}\protect\color{black}} 
\else
    \newcommand{\bae}[1]{}
    \newcommand{\di}[1]{}
    \newcommand{\ft}[1]{}
    \newcommand{\jd}[1]{}
\fi

\usepackage{lipsum}
\usepackage[colorinlistoftodos]{todonotes}

\usepackage{amsthm}

\usepackage[colorlinks = True, linkcolor = blue, citecolor = blue]{hyperref}

\DeclareMathOperator*{\argmin}{arg\,min}
\newtheorem{problem}{Problem}
\newtheorem{definition}{Definition}

\newcommand{\spot}[1]{\textcolor{magenta}{#1}}
\newcommand{\strat}[1]{\textcolor{orange}{#1}}
\newcommand{\baseline}[1]{\textcolor{Violet}{#1}}
\newcommand{\best}[1]{\textcolor{Green}{#1}}
\usepackage{svg}

\title{\LARGE \bf
Occupancy-aware Trajectory Planning for Autonomous Valet Parking in Uncertain Dynamic Environments
}

\begin{document}

\author{Farhad Nawaz$^{*1, 2}$, Faizan M. Tariq$^{1}$, Sangjae Bae$^{1}$, David Isele$^{1}$, Avinash Singh$^{1}$, \\ Nadia Figueroa$^{2}$, Nikolai Matni$^{2}$ and Jovin D'sa$^{*1}$     
\thanks{
$^{1}$Honda Research Institute (HRI), San Jose, CA 95134, USA.}
\thanks{$^{2}$GRASP Lab, University of Pennsylvania, PA 19104, USA.}
\thanks{$^{*}$Corresponding authors: farhadn@seas.upenn.edu, jovin\_dsa@honda-ri.com.  All work was done when Farhad Nawaz was employed by HRI.}
}

\maketitle


\begin{abstract}
Autonomous Valet Parking~(AVP) requires planning under partial observability, where parking spot availability evolves as dynamic agents enter and exit spots. Existing approaches either rely only on instantaneous spot availability or make static assumptions, thereby limiting foresight and adaptability. We propose an approach that estimates probability of future spot occupancy by distinguishing initially vacant and occupied spots while leveraging nearby dynamic agent motion. We propose a probabilistic estimator that integrates partial, noisy observations from a limited Field-of-View, with the evolving uncertainty of unobserved spots. Coupled with the estimator, we design a strategy planner that balances goal-directed parking maneuvers with exploratory navigation based on information gain, and incorporates wait-and-go behaviors at promising spots. Through randomized simulations emulating large parking lots, we demonstrate that our framework significantly improves parking efficiency and trajectory smoothness over existing approaches, while maintaining safety margins. 
\end{abstract}

\section{INTRODUCTION}
\label{sec:intro}
\begin{figure}[!t]
         \centering  
\includegraphics[width=\linewidth]{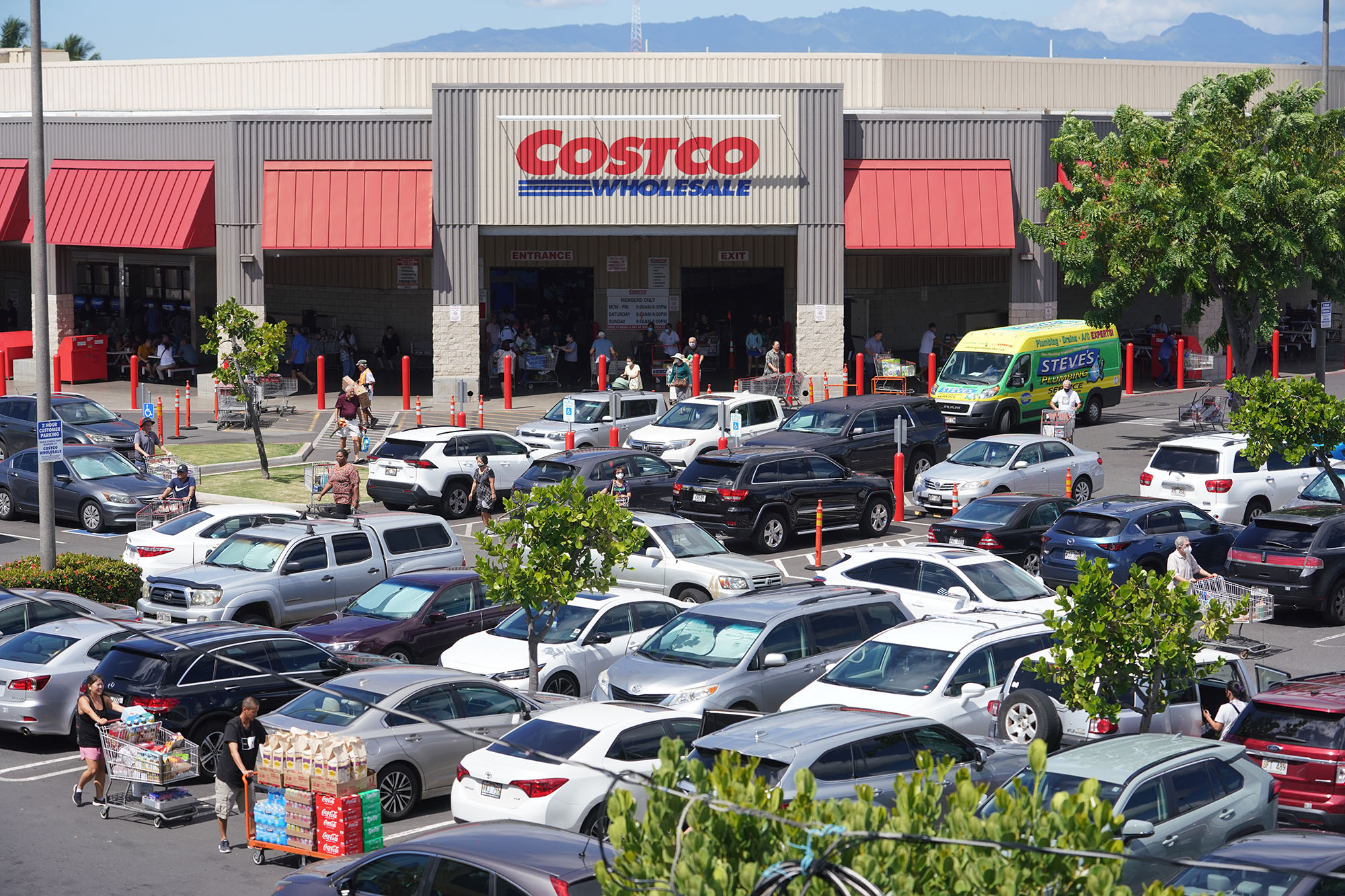}
         \vspace{-5pt}
      \caption{Example of a crowded shopping mall parking lot~\cite{costco_art}.} 
        \label{fig:costco}
\end{figure}

In busy public spaces, parking consumes significant time, space, and fuel~\cite{survey_1, survey_2}. Drivers often loop through lots such as shown in Fig.~\ref{fig:costco}, searching for spots, interacting with other drivers and pedestrians, and performing tight maneuvers. While existing park assist systems can steer a car into a user-selected spot~\cite{autopark, tesla_ref}, a Type-1 autonomous valet parking system~\cite{ISO_23374_1_2023} that operates without infrastructure support remains unavailable. To this end, we propose a framework for Type-1 AVP that integrates an observation model tailored for onboard sensing with unified spot selection and trajectory planning.

Several trajectory planners for autonomous parking have been proposed using optimization and learning-based approaches. Some optimization-based methods often focus on static obstacle avoidance~\cite{OBCA, iter_spline}, while dynamic obstacle avoidance approaches~\cite{SG_control, OBCA_dyn} typically rely on predefined reference paths. Time-optimal formulations use dynamic optimization~\cite{li2016time, li2015unified} to reduce parking time but require high computational effort, whereas heuristic planners that use Reeds–Shepp curves~\cite{kim2014auto} trade optimality for faster real-world execution. Learning-based approaches leverage deep neural networks~\cite{kim2023neural, chai2022deep, chai2020design} and reinforcement learning~\cite{multi-RL, unknown} to generate diverse trajectories in highly constrained and partially unknown environments, but fail to guarantee safety. While these approaches demonstrate effective maneuver execution, they generally assume predefined spot availability and decouple spot selection from trajectory planning.

Methods that focus on parking spot selection often make simplifying assumptions that limit practical applicability. Some approaches ignore spot occupancy prediction~\cite{game}, while others assume full connectivity between vehicles~\cite{parking_connected}. Reinforcement learning strategies~\cite{rl1_spot_assign, rl2_spot_assign} optimize parking space allocation and reduce search time, but struggle to generalize and guarantee real-time safety. Vanilla Bayesian filtering methods~\cite{inf_path} typically neglect the influence of dynamic agents on spot availability, and their observation models~\cite{inf_path, game} assume fixed sensor confidence within the Field-of-View~(FoV), overlooking the practical degradation of perception reliability with distance. Exploration-driven strategies~\cite{entropy_assign, inf_path} focus on information gain but often disregard the downstream task of actual parking. For instance,~\cite{entropy_assign} uses a simplified entropy-based occupancy model that ignores vehicle dynamics, limiting its suitability in dynamic environments. Connected-vehicle frameworks~\cite{parking_connected, fleet} assume Vehicle-to-Vehicle (V2V) and Vehicle-to-Infrastructure (V2I) communication, which are effective for offline spot allocation, but do not address the real-time strategic adjustments required in infrastructure-free scenarios.

To address these gaps, we propose an occupancy-aware trajectory planning framework tailored for Type-1 AVP:
\begin{itemize}
    \item \textbf{Distance-aware partial FoV model}: We model partial observability by introducing a distance-dependent FoV model that maps range to observation confidence. 
    \item \textbf{Spot occupancy estimator:}  We propose a Bayes filter based approach that predicts the future occupancy of parking spots using the behavior of dynamic agents. We also model distinct arrival and departure equations for initially vacant and occupied spots, respectively, to reason about the future occupancy.
    \item \textbf{Strategy planner:} We integrate future spot occupancy probabilities into a cost-based policy that plans the ego vehicle’s trajectory to either explore the parking lot or commit to a spot. The policy allows the ego to pause near promising spots by incorporating a waiting-time cost, enabling it to capture imminent vacancies while maintaining a balance between efficiency and safety.
\end{itemize}
We evaluate our approach in a simulated parking lot, combining foresighted occupancy reasoning and strategy planning. Our method reduces time and distance traveled while generating smoother and safer paths compared to existing spot selection methods and trajectory planners~\cite{inf_path, parking_connected}.


\section{Problem formulation and Approach}
\label{sec:prob_form}

We consider an ego vehicle navigating a parking lot, equipped with onboard sensors that provide observations of spot occupancy within a limited FoV, along with motion predictions of dynamic agents such as vehicles and pedestrians. The task for the ego is to plan trajectories that efficiently explore the lot, identify an available spot, and park safely.

\subsection{Parking Lot Environment}

\begin{figure}[!b]
         \centering  
\includegraphics[width=\linewidth]{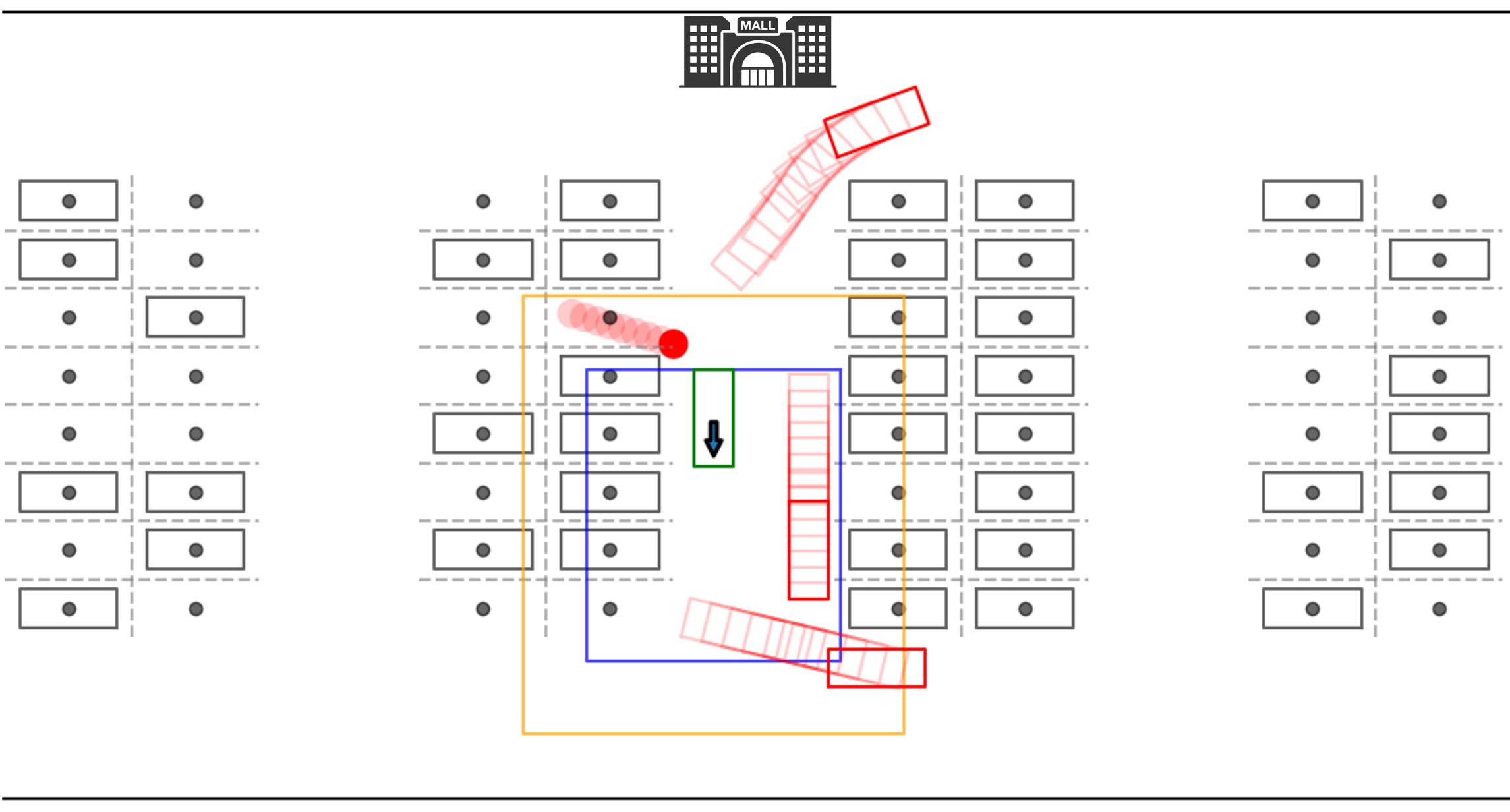}
         \vspace{-5pt}
      \caption{Autonomous ego vehicle (green) with limited field of view (blue/orange) navigating an environment simulating a mall parking lot, with the entrance at the top. Predicted motions of dynamic vehicles are shown as red rectangles, and a pedestrian is indicated by a red circle. Spots are fully observed inside blue rectangle~($\epsilon=1$), unobserved outside orange~($\gamma=1.5$), and partially observed in between. Static vehicles are black, and parking spot centers are grey dots. } 
        \label{fig:prob_scenario}
\end{figure}

Consider the parking lot in Fig.~\ref{fig:prob_scenario} with $N_p$ spots, whose centers are given by $\mathcal{P} = \{s_1, \dots, s_{N_p}\}$, where $s_i$ denotes the 2D coordinates of spot~$i$. The ego vehicle has prior access to a static map of the lot (boundaries and lane centerlines/widths), but spot occupancy is unknown and must be observed in real time. Parking lot boundaries are treated as static obstacles as given in~\cite{farhad_IV}. 


\subsection{Vehicle Model}
Let $x = [X, Y, \theta]^\top$ be the state of the vehicle, where $(X, Y)$ is the center of the rear axis and $\theta$ is the heading angle. We model vehicle dynamics using the kinematic bicycle model~\cite{rajamani2011vehicle}, which is well suited for low-speed maneuvers:
\begin{equation}
\dot{x} = f(x, u) \Leftrightarrow \begin{bmatrix} \dot{X} \\ \dot{Y} \\ \dot{\theta} \end{bmatrix} = \begin{bmatrix}
v \cos(\theta) \\ v \sin(\theta) \\ \frac{v}{L}\tan(\delta)
\end{bmatrix} .
    \label{dynamics}
\end{equation}
The control input is $u = \begin{bmatrix}
    v \\ \delta
\end{bmatrix}$, where $v$ and $\delta$ is the longitudinal velocity and steering angle of the front wheel, respectively. The wheelbase of the vehicle is $L$, and the vehicle is modeled as a rectangle.

\vspace{-2pt}
\subsection{Observation Model} 
\label{sec:obs_model}
\vspace{-2pt}
The vehicle’s limited onboard sensing is modeled via a “scaled and shifted” infinity norm defining the FoV, though other shapes (e.g., ellipses or circles) can be modeled by modifying the \textit{distance metric}. This formulation captures the limited and imperfect nature of real-world observations.

\begin{definition}[Distance metric]
\textnormal{
The scaled and shifted distance between the ego vehicle at state~$x= \begin{bmatrix}
    X&Y&\theta
\end{bmatrix}^{\top}$ and a point~$s \in \mathbb{R}^2$ is defined as
\begin{align}
    d_{(\zeta, r)}(x, s) &= \left\|r^{\top} R(\theta)^{\top} \left(s - \left(\begin{bmatrix}
        X \\ Y
    \end{bmatrix} + R(\theta)\begin{bmatrix}\zeta \\ 0\end{bmatrix}\right)\right) \right\|_{\infty}, \nonumber \\
    \textnormal{such that} \ r &= \begin{bmatrix} \frac{1}{r_x} & \frac{1}{r_y} \end{bmatrix}^{\top}, \ R(\theta) = \begin{bmatrix}
        \cos(\theta) & -\sin(\theta) \\
        \sin(\theta) & \cos(\theta)
    \end{bmatrix},
 \label{inf_scaled_shifted}
\end{align}
where $r_x$ and $r_y$ are the half-lengths of the FoV rectangle along the longitudinal and lateral directions, respectively, while $\zeta$ is the forward shift in ego frame to bias the sensing region ahead of the vehicle. The rotation matrix $R(\theta)$ transforms vectors from the ego frame to the world frame. }
    \label{def:dist}
\end{definition}

\begin{definition}[Spot observability]
\textnormal{
The \textit{observability} of a spot $s \in \mathcal{P}$ from vehicle state $x$ is defined as:
\begin{align}
    \text{Spot}~s \text{ is}
&\begin{cases}
\text{fully observed if}  &d_{(\zeta, r)}(x, s) \leq \epsilon, \\
\text{unobserved if}  &d_{(\zeta, r)}(x, s) \geq \gamma, \\
\text{partially observed} &\text{otherwise},
\end{cases}
\end{align}
where $\epsilon < \gamma$ and are obtained from the vehicle sensing limits. 
}
\label{def:observability}
\end{definition}
\vspace{-15pt}
The different observability regions are illustrated in Fig.~\ref{fig:prob_scenario}.
\vspace{-15pt}
\begin{definition}[Observation accuracy]
\textnormal{
Let $O_t^s \in \{0,1\}$ be the ground-truth occupancy and $Z_t^s \in \{0,1\}$ be the observed occupancy of spot $s$ at time $t$, where $1 :=$ occupied and $0 :=$ vacant. Observation accuracy is the likelihood of correctly observing spot $s$, modeled using the distance-dependent probability 
\vspace{-5pt}
\begin{equation}
\begin{aligned}
& p_c(d_{(\zeta, r)}(x, s)) = P(Z_t^s = 1 | O_t^s = 1) = P(Z_t^s = 0 | O_t^s = 0), \\
& p_c(d_{(\zeta, r)}(x, s))  = e^{-\ln(2)\left(1 + e^{-\alpha_c \left(d_{(\zeta, r)}(x, s) - \left(\frac{\gamma - \epsilon}{2} + \epsilon\right)\right)}\right)^{-1}}, 
\end{aligned}
\label{prob_correct_dist}
\end{equation}
where~$p_c(d)$ is the exponential of a scaled sigmoid centered at $\frac{\gamma - \epsilon}{2} + \epsilon$, such that $p_c(d) \to 1$ as $d$ decreases to $\epsilon$ and $p_c(d)$ decreases to $0.5$ as $d \to \gamma$.The probability~\eqref{prob_correct_dist} models observation confidence degrading with distance from ego vehicle, unlike prior work~\cite{inf_path} that assumes fixed uncertainty. 
}
\label{def:obs_acc}
\end{definition}
The formal problem statement is given below.
\begin{problem}
\textnormal{
Given the ego vehicle state~$x_t \in \mathbb{R}^3$ at time $t$, parking spot observations~$Z_t^s$ as defined in Section~\ref{sec:obs_model}, predictions of $D$ dynamic agents~$\{\{\hat{y}^j(k)\}_{k=t}^{t+T}\}_{j=1}^{D}$ over a horizon~$T$, where $\hat{y}^j(k) \in \mathbb{R}^3$ is the position and heading of agent~$j$ if it is a vehicle, and $\hat{y}^j(k) \in \mathbb{R}^2$ is the position if agent~$j$ is a pedestrian; compute a reference trajectory that satisfies the vehicle dynamics~\eqref{dynamics} and steers the vehicle toward a feasible parking spot or explore the lot while avoiding static and dynamic obstacles.
}
    \label{prob_state}
\end{problem}
\vspace{-5pt}
\textbf{Remark:} We assume the availability of predicted trajectories of dynamic agents over a horizon~$T$ at each planning step~$t$ to solve Problem~\ref{prob_state}, while aiming to develop and integrate a trajectory prediction model in future work.


\subsection{Proposed Approach}
\begin{figure}[!b]
         \centering         \includegraphics[width=\linewidth]{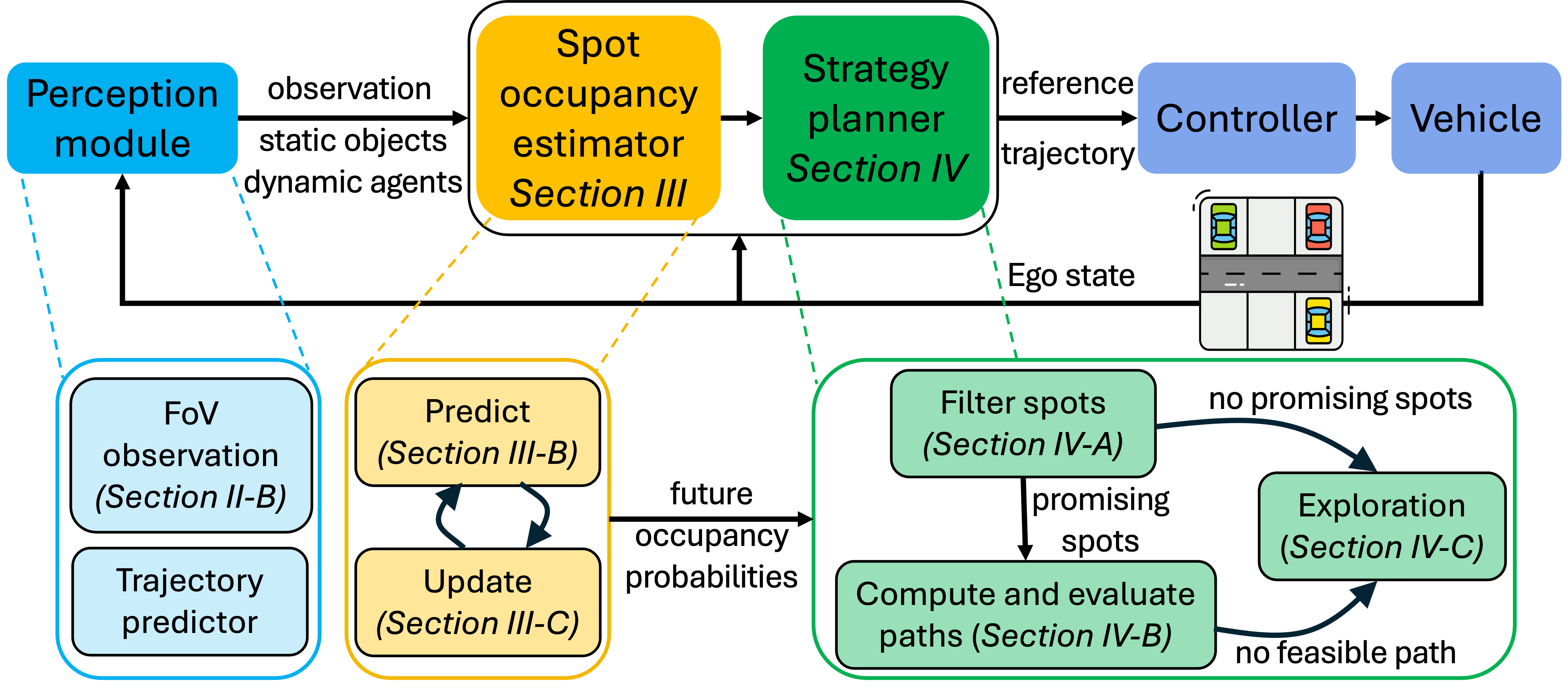}
      \caption{Outline of our proposed approach.} 
        \label{fig:approach}
\end{figure}

Fig.~\ref{fig:approach} illustrates our approach, which integrates spot occupancy estimation with planning. The perception module provides dynamic agent predictions and occupancy observations using our partial FoV model, and the occupancy estimator recursively updates to predict future spot availability. The strategy planner then computes and evaluates feasible paths, corresponding to one of the three actions—\textit{park immediately}, \textit{wait near a potential spot}, or \textit{explore the lot}—by minimizing a cost that trades off efficiency and safety. The chosen trajectory is executed by the vehicle controller, closing the loop.

\section{Spot Occupancy Estimator}
\label{sec:spot_occ_est}
At each planning step $t$, the spot occupancy estimator predicts occupancy probabilities for all spots in the lot over a horizon $T$, combining current observations with predicted motion of dynamic agents. This enables the ego vehicle to plan based on anticipated occupancies rather than instantaneous observations. Probabilities are computed via a Bayes filter~\cite{prob_robotics}, with the \textit{prediction} step modeling agent interactions and the \textit{update} step incorporating FoV observations.

\subsection{Effect of Dynamic Agents}
Each agent \(j\)’s position at time~$k$ is modeled as a Gaussian:
\vspace{-10pt}
\begin{equation}
y^j(k) \sim \mathcal{N}(\hat{y}^j(k), \Sigma(k)), \quad k \in \{t, \dots, t+T\},
\label{Gauss_pred}
\end{equation}
where~\(\hat{y}^j(k)\) is the nominal prediction and $\Sigma(k)$ is the covariance. These uncertain trajectories define the probability that a spot~$s \in \mathcal{P}$ is occupied by an agent~$j$ at time \(k\):
\begin{equation}
p(s,j,k) = E_d \cdot A_f,
\label{prob_agent_spot_time}
\end{equation}
where
\vspace{-10pt}
\begin{equation}
E_d = \big((\alpha_d+1)(\alpha_d+e^{\alpha_1 m_d(y^j(k),s)})\big)^{-1}
\label{Exp_decay}
\end{equation}
captures spatial proximity via the Mahalanobis distance~\cite{mahala}:
    \[
    m_d(y^j(k),s) = (\hat{y}^j(k)-s)^T \Sigma(k)^{-1} (\hat{y}^j(k)-s)
    \]
    and
    \vspace{-5pt}
\begin{equation}
    A_f = \left(1 + e^{-\alpha_2 \sigma_v(k)^{-1} \hat{v}^j(k)^T (s - \hat{y}^j(k))}\right)^{-1}
    \label{align_fact}
\end{equation}
is an alignment factor that reflects whether the agent is moving toward the spot, based on its predicted velocity \(\hat{v}^j(k)\) and velocity uncertainty \(\sigma_v(k)\). Together, $E_d$ and $A_f$ encode the intuition that an agent is more likely to occupy a spot if the agent is both closer and moving towards the spot, while accounting for motion uncertainty.

\textbf{Uncertainty Propagation:} We propagate the positional covariance at each timestep using the discrete-time  model:
\begin{equation}
    \Sigma(k+1) = R(\theta(k))\left(\Sigma(k) + Q\right)R(\theta(k))^{\top},
\label{unc_discrete}
\end{equation}
where $R(\theta(k))$ is the rotation matrix, $\theta(k)$ is vehicle’s heading and $Q$ is a diagonal matrix representing process noise in ego frame. The initial covariance is $\Sigma(t) = \begin{bmatrix} \beta_l L & 0 \\ 0 & \beta_w W\end{bmatrix}$, with $L$ and $W$ denoting the length and width of the vehicle, while $\beta_l$ and $\beta_w$ correspond to the respective scaling factors. If the dynamic agent is a pedestrian, then $\theta(k) = 0 \deg$, $\beta_l = \beta_w = 1$, and we use a radius parameter to represent the size of the pedestrian instead of the length and width of the vehicle. To estimate the uncertainty in velocity, we approximate the instantaneous velocity as the finite difference $v^j(k) = \frac{y^j(k+1) - y^j(k)}{\Delta t}$, where $\Delta t$ is the time interval between successive time steps. The resulting covariance in velocity is
\[
\Sigma_v(k) = \frac{\Sigma(k+1) + \Sigma(k) - 2\Sigma_{\rho}}{\Delta t^2},
\]
where $\Sigma_{\rho} = \textnormal{Cov}\left(y^j(k+1), y^j(k)\right)$ models the covariance between consecutive position estimates. We assume a symmetric form and set $\Sigma_{\rho} = \begin{bmatrix} \rho & 0 \\ 0 & \rho\end{bmatrix}$ for all $j\in\{1,2,\ldots,D\}$ and $k\in\{t,t+1,\ldots,t+T-1\}$. Note that when $\rho \neq 0$, the velocity estimate is not Gaussian since
sum of two Gaussian random variables is a Gaussian only if they are independent (uncorrelated) or jointly Gaussian.

From~\eqref{prob_agent_spot_time}, the probability that at least one of the $D$ dynamic agents will occupy spot \(s\) at time \(k\) is:
\vspace{-5pt}
\begin{equation}
    q(s,k) = 1 - \prod_{j=1}^{D} \left(1 - p(s,j,k)\right).
    \label{prob_spot_time}
\end{equation}
At each step \(k \in \{t, \dots, t+T\}\), we predict the prior belief \(\bar{b}_k^s = P(O_k^s = 1)\) using~\eqref{prob_spot_time}, and update it with the observations~\(Z_k^s\) to obtain the posterior belief~\({b_k^s = P(O_k^s = 1 \mid Z_k^s)}\), as described in Section~\ref{sec:pred}.

\subsection{Prediction}
\label{sec:pred}

Prior work~\cite{inf_path} propagates beliefs without considering the effect of dynamic agents on parking spots. We explicitly incorporate dynamic agent motion and model asymmetric interaction with spots: the ego cares about a currently occupied spot becoming vacant, while it is concerned about a vacant spot becoming occupied in the future. This distinction enables faster parking and smoother trajectories, as shown in Section~\ref{sec:exp_spot_est}. The current occupancy of a spot is determined from its observation~$Z_t^s$. No occupancy observations and agent interactions are available outside the FoV.
\begin{enumerate}
    \item For initially \textbf{vacant} spots, our transition model emphasizes incoming occupancy, and the prior belief is recursively updated as
    \begin{equation}
    \bar{b}_{k+1}^s =  q(s, k+1) (1 - b_k^s) + \mu_1 b_k^s,
        \label{incoming_predict}
    \end{equation}
    where $q(s, k+1) = P(O_{k+1}^s = 1 | O_{k}^s = 0)$ is computed from~\eqref{prob_spot_time} and $\mu_1 =  P(O_{k+1}^s = 1 | O_{k}^s = 1)$ models the departure probability of a parked vehicle using an exponential distribution~\cite{parking_lot_data}, reflecting the memory-less nature of departures. If \(\lambda_d\) denotes the average departure rate of a parked vehicle, the probability of departure within a time interval \(\Delta t\) is
    \begin{equation}
        P(O_{k+1}^s = 0 | O_{k}^s = 1) = 1-e^{-\lambda_d \Delta t} \Rightarrow \mu_1 = e^{-\lambda_d \Delta t}.    
        \label{dep_rate}
    \end{equation}
    \item For initially \textbf{occupied} spots, the emphasis is on potential departure of the occupied agent, captured as
    \begin{equation}
    1-\bar{b}_{k+1}^s =  \mu_2 (1 - b_{k}^s) + (1-q(s, k+1)) b_k^s,
        \label{outgoing_predict}
    \end{equation}
    where $1-q(s, k+1) = P(O_{k+1}^s = 0 | O_{k}^s = 1)$ is computed from~\eqref{prob_spot_time} and $\mu_2 =  P(O_{k+1}^s = 1 | O_{k}^s = 0)$ models the arrival probability of a vacant spot using a Poisson distribution, which commonly describes arrival processes~\cite{arrival_mark}. If \(\lambda_a\) denotes the average number of vehicles that arrive with the time interval~$\Delta t$, the probability that no vehicle parks in a vacant spot is
    \begin{equation}
    \begin{aligned}
        P(O_{k+1}^s = 0 | O_{k}^s = 0) = e^{-\lambda_a} \Rightarrow \mu_2 = 1 - e^{-\lambda_a}.
    \end{aligned}      
    \label{arr_rate}
    \end{equation}
    \item  For \textbf{unobservable} spots, the prediction equation depends only on the average departure and arrival rates:  
    \vspace{-10pt}
\begin{equation}
\bar{b}_{k+1}^s =  \mu_2 (1 - b_{k}^s) + \mu_1 b_k^s.
    \label{unobserved_predict}
\end{equation}
\end{enumerate}
\vspace{-5pt}
The prior~$\bar{b}_{k+1}^s$ serves as the input to the update step, where it is fused with observations.

\subsection{Update}

The update combines current spot observations with~$\bar{b}_{k+1}^s$ to estimate future occupancy. For fully observed spots, i.e., $p_c(d_{(\zeta, r)}(x_t, s)) = 1$, we completely trust the current observation and propagate occupancy using only agent predictions. For partially observed or unobserved spots, i.e., $p_c(d_{(\zeta, r)}(x_t, s)) < 1$, we apply Bayes' rule to fuse the observation with the prior belief:
\vspace{-5pt}
\begin{equation}
b_{k+1}^s = \begin{cases}
\bar{b}_{k+1}^s & \textnormal{if} \ p_c(d_{(\zeta, r)}(x_t, s)) = 1\\
    \frac{P(Z_{k+1}^s | O_{k+1}^s = 1) \bar{b}_{k+1}^s}{P(Z_{k+1}^s)} & \textnormal{otherwise},
\end{cases}
    \label{update}
\end{equation}
where $x_t$ is the current ego state. Since only the current occupancy observation~$Z_t^s$ is available, we set $Z_{k+1}^s = Z_t^s$ for all $k\in\{t,\ldots,t+T-1\}$. For unobserved spots~$s$, from~\eqref{prob_correct_dist}, we have~${p_c(d_{(\zeta, r)}(x_t, s)) \approx 0.5}$. Hence, using~\eqref{update}, the update equation reduces to~$b_{k+1}^s\approx \bar{b}_{k+1}^s$, i.e., the prior belief remains unchanged.


\section{Strategy Planner}
\label{sec:strat_planner}

Given the probabilistic forecasts from the spot occupancy estimator (Sec.~\ref{sec:spot_occ_est}), the strategy planner selects actions (park, wait, or explore) and generates a reference trajectory that balances efficiency and safety in dynamic parking scenarios.

\subsection{Filter Spots}

At each planning step, we filter candidate spots inside the FoV based on the occupancy predictions, considering two cases: (i) initially vacant and remain vacant (ii) initially occupied and become vacant. Formally, spot $s$ is vacant in the future (t+T) if:
\vspace{-5pt}
\begin{align}
&\begin{cases}
s~\text{is \textbf{initially vacant} (time } t) \text{ and } b_{t+T}^s \leq P_v, \\
s~\text{is \textbf{initially occupied} (time } t) \text{ and } b_{t+T}^s \leq P_o.
\end{cases}
    \label{filter_future}
\end{align}
This filtering ensures that the strategy planner only considers feasible parking spots, enhancing decision-making efficiency in dynamic environments. We set a spot is more likely to be occupied if initially vacant~($P_v < P_o$) and vice-versa to reflect the asymmetry in availability reasoning.


\subsection{Generate and Evaluate Paths}

At planning step~$t$, let $\mathcal{S}_t \subset \mathcal{P}$ denote spots classified as vacant by~\eqref{filter_future}. We generate trajectories ${\mathcal{T} = \{\tau_g\}_{g \in \mathcal{S}_t}}$ using Hybrid A$^\star$\footnote{Alternatives include sampling or spline-based planners~\cite{mppi, recurr_spline, iter_spline}.}, where each $\tau_g$ leads to a goal state $g$ from the ego state~$x_t$. Each trajectory is defined as ${\tau_{g} = \{x(k)\}_{k=t}^{t + N_{g}}}$, where ${x(t) = x_t}$, $x(t + N_{g}) = g$ and $N_{g}$ is the time required to reach~$g$ from~$x_t$. While static obstacles are considered in trajectory generation, dynamic obstacles are handled via the cost function and waiting time parameter, defined in the next section. Trajectory generations and evaluations are parallelized across all goal states~$g$ for efficiency. 

\subsubsection{Cost function} 
The cost function primarily evaluates efficiency (in time and distance), safety (w.r.t. collision avoidance), and smoothness (w.r.t. acceleration and gear changes):
\vspace{-5pt}
\begin{align}
&\mathcal{C}(\tau_{g}) = N_{g} + t_w +  \sum_{k=t}^{t+t_w } c_o(x(k), \mathcal{Y}(k)) +  \nonumber \\ 
&\sum_{k=t + t_w+1}^{t + t_w + N_{g}} c_o(x(k-t_w), \mathcal{Y}(k)) + \sum_{k=t}^{t + N_{g}-1} c_{\textnormal{smooth}}(x(k)),
    \label{wait_go_cost}
\end{align}
\vspace{-5pt}
where
\vspace{-5pt}
\begin{align}
    c_{o}(x(k), \mathcal{Y}(k)) &= \sum_{j=1}^{D} h(x(k), \hat{y}^j(k)) 
    + \sum_{i=1}^{W} h(x(t), w_i), \\
    c_{\textnormal{smooth}}(x(k)) &= \|a(k)\| + \textrm{D}_\text{C}(v(k),v(k+1)) + \nonumber \\ 
    &\quad \mathds{1}(\textrm{reverse}(k)) 
    + \mathds{1}(\textrm{change\_gear}(k)),
    \label{cost_path}
\end{align}
$t_w$ denotes the \textit{waiting time} (explained in the next section), ${\mathcal{Y}(k) = \{\hat{y}^j(k)\}_{j=1}^{D}}$ represents the configuration of dynamic agents at time~$k$ and the static obstacles are given by $\{w_i\}_{i=1}^{W}$ with $w_i \in \mathbb{R}^2$. At each time step, $c_o(\cdot, \cdot)$ penalizes proximity to obstacles and
$c_{\textnormal{smooth}}(\cdot)$ promotes smooth trajectories. 

In $c_o(\cdot,\cdot)$, the obstacle avoidance cost~$h(\cdot, \cdot)$ is
\begin{align}
h(x, y) = e^{-\alpha_o d_o(x, y)},
\label{obst_avoid}
\end{align}
where $d_o(x, y)$ is the distance between the ego vehicle~$x$ and obstacle~$y$. The barrier cost~$h(\cdot, \cdot)$ penalizes proximity to obstacles, with $\alpha_o = 2$ for dynamic agents and $\alpha_o = 3$ for static obstacles, reflecting lower penalties for static ones. We extend the 3-circle distance from~\cite{nnmpc} to a 5-circle model for $d_o(\cdot, \cdot)$ to improve accuracy in dynamic collision avoidance. For static obstacles, the exact distance between the ego vehicle’s edge and obstacle points~$\{w_i\}_{i=1}^W$ is computed as given in Section III of~\cite{farhad_IV}, since static obstacles are more reliably perceived. To balance terms, $\|a(t)\|$ and the cosine distance are scaled to $[0,1]$ using velocity limits. 

In~\eqref{smooth_cost}, $\text{D}_\text{C}(\cdot, \cdot)$ is the cosine distance~\cite{cosinedistance} and
\begin{align}
&a(k) = \begin{bmatrix}
    \ddot{X}(k), \ddot{Y}(k)
\end{bmatrix}^{\top}, \ v(k) = \begin{bmatrix}
    \dot{X}(k), \dot{Y}(k)\end{bmatrix}^{\top}, \nonumber \\
    &\mathds{1}(\textrm{reverse}(k)) = \begin{cases}
        1, & \textrm{if} \ \textrm{D}_\text{C}\left(v(k),\begin{bmatrix}
            \cos(\theta(k)) \\ \sin(\theta(k))
        \end{bmatrix}\right) > 1,\\
        0, & \textrm{otherwise}         
    \end{cases}, \nonumber\\    
    &\mathds{1}(\textrm{change\_gear}(k)) = \begin{cases}
        1, & \textrm{if} \ \textrm{D}_\text{C}\left(v(k),v(k+1)\right) > 1,\\
        0, & \textrm{otherwise}.      
    \end{cases}
    \label{smooth_cost}
\end{align}

\vspace{-4pt}
\subsubsection{Waiting time} 
We incorporate \textit{wait-and-go} behaviors in the trajectories~$\tau_g$ by introducing a waiting time $t_w$ in~\eqref{wait_go_cost}. This enables the ego vehicle to pause and yield to dynamic agents before moving to park in a spot~$g$, thereby reducing collision risks in constrained parking lots. To ensure feasibility, we impose a maximum wait time $T_w$ and consider only paths with $t_w \leq T_w$. The value of $T_w$ is set based on parking lot density, with future work aimed at dynamically adapting it to congestion.

\textbf{Remark:} If $T<N_{g}$, we extrapolate the predictions using a constant velocity model: ${\hat{y}^j(k) = \int_{s=t+T}^k \hat{v}^j(t+T-1) ds}$ for all ${k \geq t+T+1}$ and ${j \in \{1,2,\ldots,D\}}$, where ${\hat{v}^j(k) = \frac{\hat{y}^j(k+1) - \hat{y}^j(k)}{\Delta t}}$.

Once we compute and evaluate the trajectories~$\mathcal{T}$, we choose the one that has the minimum cost~$\argmin_{\tau \in \mathcal{T}}\mathcal{C}(\tau)$ and avoids all obstacles respecting the safety thresholds. We set larger safety thresholds for dynamic agents—$50$cm for vehicles and $90$cm for pedestrians—compared to $20$cm for static obstacles. These values are tailored to parking scenarios but can be adjusted for other applications.

\vspace{-3pt}
\subsection{Exploration}

If no feasible trajectory exists to a spot in $\mathcal{S}_t$, the ego vehicle switches to exploratory mode, selecting a trajectory that maximizes information gain about spot occupancy. In typical parking lots like Fig.~\ref{fig:prob_scenario}, maneuver options are limited to continuing straight within a row, entering a row from outside, or turning left or right at the end. We generalize this by evaluating paths to a finite set of exploration goals $\mathcal{E} \subset \mathbb{R}^3$, where each $e \in \mathcal{E}$ represents a candidate goal pose.

\subsubsection{Exploration goals}
We define three exploration goals~${\mathcal{E} = \{e_1, e_2, e_3\}}$ at each time step: going straight ($e_1$), turning left ($e_2$), and turning right ($e_3$). These options suit structured parking lots with parallel rows (Fig.~\ref{fig:prob_scenario}), though we plan to extend them to more general layouts such as multi-level garages in future work. The straight goal is:
\begin{equation}
e_1 = \begin{bmatrix}
    R(\theta_t)\begin{bmatrix}
    \epsilon - \eta
    \\ 0
\end{bmatrix} + \begin{bmatrix}
    X_t
    \\ Y_t
\end{bmatrix} \\ \\
\theta_t
\end{bmatrix},
\label{go_straight}
\end{equation}
where $x_t = [X_t, Y_t, \theta_t]^{\top}$ is the ego state, $\epsilon$ is the full observability threshold (Sec.~\ref{sec:obs_model}), and $\eta \in [0, \epsilon)$ denotes how far the ego moves forward. Left and right turn goal states for right-hand traffic convention are described as
\begin{equation}
e_\sigma =
\begin{bmatrix}
R(\theta_t)
\begin{bmatrix}
x_{\text{road}} + \sigma \cdot \left(\frac{l_w}{2}\right) \\ \sigma \cdot \epsilon
\end{bmatrix}
+ 
\begin{bmatrix} X_t \\ Y_t \end{bmatrix} \\
\theta_t + \sigma \cdot \left(\frac{\pi}{2}\right)
\end{bmatrix}, \quad \sigma\in\{+1, -1\},
\label{left_right}
\end{equation}
where~$e_2 = e_{+1}$(left), $e_3 = e_{-1}$(right), $x_{\textnormal{road}}$ is the center of the newly observed road, and $l_w$ its lane width.

\subsubsection{Information gain}
We use Hybrid A$^\star$ to generate trajectories to $e_2$ and $e_3$, and a 5\textsuperscript{th}-order spline to interpolate from $x_t$ to $e_1$. Let $\mathcal{T}' = \{\tau_e\}_{e \in \mathcal{E}}$ denote the set of trajectories to each exploration goal. The information gain of a trajectory is evaluated using the entropy $\mathcal{H}(\cdot)$ of a spot $s$ at time $k$:
\vspace{-5pt}
\begin{equation}
\mathcal{H}(b_k^s) = -b_k^s \log_2(b_k^s) - (1-b_k^s)\log_2(1 - b_k^s),
    \label{entropy}
\end{equation}
where $b_k^s = P(O_k^s = 1 | Z_k^s)$. We simulate future observations $\{Z_k^s\}_{k=t+1}^{t+T}$ for each trajectory~$\tau_e$ and denote~$Z_k^s(\tau_e)$ to be the observation of spot~$s$ at time~$k$ when the ego follows~$\tau_e$. Then, we recursively update beliefs~$b_k^s(\tau_e)$ using the update rule~\eqref{update} and the simulated observations~$Z_k^s(\tau_e)$ for each~$\tau_e$. The total information gain of a trajectory is defined using the entropy reduction of all parking spots when following~$\tau_e$:
\vspace{-5pt}
\begin{equation}
\mathcal{I}(\tau_e) = \sum_{i=1}^{N_p} \left(\mathcal{H}(b_t^{s_i}(\tau_e)) - \mathcal{H}(b_{t+N_e}^{s_i}(\tau_e))\right),
    \label{inf_gain}
\end{equation}
where $N_e$ is the length of~$\tau_e = \{x(k)\}_{k=t}^{t+N_e}$. To balance exploration with motion efficiency, we use~\eqref{wait_go_cost} and~\eqref{inf_gain} to define the exploration cost
\vspace{-5pt}
\begin{equation}
\mathcal{C}_e(\tau_e) = \mathcal{C}(\tau_e) - \mathcal{I}(\tau_e),
\label{cost_exp}
\end{equation}
and select the optimal trajectory as \( \argmin_{\tau_e \in \mathcal{T}'} \mathcal{C}_e(\tau_e) \), subject to safety thresholds. If no safe trajectory exists, a contingency planner keeps the vehicle stationary or steers it away from adversarial dynamic agents on a collision course.

\section{Experimental Results}

\begin{figure}[!b]
        \centering
\includegraphics[width=\linewidth]{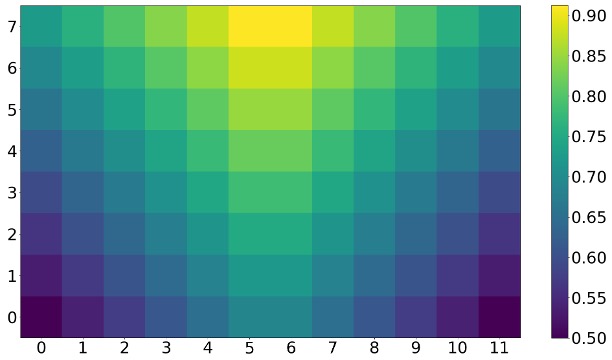}
      \caption{Initial occupancy probabilities for the parking lot in Fig.~\ref{fig:prob_scenario}, consisting of 8 rows and 12 columns of spots.} 
        \label{fig:exp_oc_prob}
\end{figure}

The ego vehicle’s task is to navigate from its initial position to a feasible parking spot, assuming real-time onboard sensor data on static vehicles, driving aisles, and dynamic agents within the FoV. We evaluate our framework in a simulated shopping mall parking lot~(Fig.~\ref{fig:prob_scenario}), with randomized initial spot occupancy via a binomial distribution reflecting higher congestion near the entrance~(Fig.~\ref{fig:exp_oc_prob}). We conduct ablation studies and comparisons with existing valet parking methods that are summarized in Table~\ref{table:metrics}, with 50 randomized runs per method varying initial occupancy, ego start position, and dynamic agent behavior. The ego is initialized either inside a parking row or outside, while other vehicles may start in vacant spots or outside, with their paths generated by a local Hybrid A$^\star$  planner for exiting or entering spots, respectively. Dynamic pedestrians are randomly spawned outside the occupied spots with velocities $v \sim \mathcal{U}[-1.5, 1.5]$ m/s.  All simulations are run in Python 3.13 on Ubuntu 20.04 with an Intel Xeon E5-2643 v4 CPU. 

\vspace{1pt}

\textbf{Simulation parameters:} The vehicle dynamics are modeled using the kinematic bicycle model~\cite{rajamani2011vehicle}, discretized with a time step of $0.1$ s for simulation. The vehicle parameters are listed in Table~\ref{tab:veh_params} based on the 2024 Honda Accord~\cite{honda_acc}. Our Hybrid A$^\star$ planner employs a state grid resolution of $0.2$ m in both $X$ and $Y$ directions, and $10 \ \textrm{deg}$ for heading. Steering and velocity inputs are each discretized into 5 values over their limits $[-\delta_{\textrm{max}}, \delta_{\textrm{max}}]$ and $[-v_{\textrm{max}}, v_{\textrm{max}}]$, respectively. Parking lot parameters in Table~\ref{tab:park_params} are adopted from~\cite{park_dim}. The prediction horizon is $T=5$ s with a discretization step of~$0.1$ s, and the maximum wait time is~$T_w=5$ s. The average arrival and departure rates are ~$\lambda_a= 0.000624~\textrm{s}^{-1}$~\eqref{arr_rate} and~$\lambda_d=0.000378~\textrm{s}^{-1}$~\eqref{dep_rate}, respectively, obtained from~\cite{parking_lot_data}. The probability thresholds in~\eqref{filter_future} are $P_v = 0.3$ and $P_o = 0.7$. The scaling factor is~$r = \begin{bmatrix} (2.5V_L)^{-1} & (3V_W)^{-1} \end{bmatrix}^{\top}$ and forward shift is~$\zeta = 0.5V_L$ in~\eqref{inf_scaled_shifted}, while $\eta = 0.5~\textrm{m}$ in~\eqref{go_straight}. The observability thresholds are~$\epsilon = 1 $ and~$\gamma = 1.5$, and $\alpha_c = 25$ in~\eqref{prob_correct_dist}.

\vspace{-5pt}
    \begin{minipage}[t]{0.45\linewidth}
        \centering
        \captionof{table}{Vehicle parameters}
        \vspace{-5pt}
        \label{tab:veh_params}
        \begin{tabular}{|p{2.6cm}|c|}
            \hline
            \textbf{Parameter} & \textbf{Value} \\
            \hline
        Length $V_L$ & 4.97 m \\
        \hline
        Width $V_W$ & 1.86 m \\
        \hline
        Wheelbase $L$ & 2.83 m \\
        \hline
        Speed limit $v_{\textrm{max}}$ & 3.5 m/s \\
        \hline
        Steering limit $\delta_{\textrm{max}}$ & 34.9 deg \\
            \hline
        \end{tabular}
    \end{minipage}%
    \hspace{22pt}
    \begin{minipage}[t]{0.35\linewidth}
        \centering
        \captionof{table}{Parking lot parameters}
        \vspace{-5pt}
        \label{tab:park_params}
        \begin{tabular}{|p{1.6cm}|c|}
            \hline
            \textbf{Parameter} & \textbf{Value} \\
            \hline
        Spot length & 6.1 m \\
        \hline
        Spot width & 2.74 m \\
        \hline
        Road width~$l_w$ & 7.62 m \\
            \hline
        \end{tabular}
    \end{minipage}


\subsection{Ablation Studies}
\label{sec:abl}

We assess our contributions by performing ablation studies on the spot occupancy estimator and strategy planner. In each case, we fix one module and vary components of the other, comparing the variants with our complete approach.

\subsubsection{Spot Occupancy Estimator}
\label{sec:exp_spot_est}

We fix our strategy planner~(Section~\ref{sec:strat_planner}) and compare the performance of three different spot occupancy estimators with ours~(Section~\ref{sec:spot_occ_est}).
\begin{itemize}
    \item \textbf{Greedy} approach assumes a horizon of $T=0.1~\textrm{s}$~(one step estimation) in Section~\ref{sec:spot_occ_est}, primarily relying on the current occupancy probabilities. 
    \item \textbf{Identical prediction} method applies the same prediction equation~\eqref{incoming_predict} regardless of whether a spot is initially vacant or occupied. 
    \item \textbf{Position only} approach models dynamic agents without incorporating velocity by removing the alignment factor~\eqref{align_fact} in~\eqref{prob_agent_spot_time} so that~$p(s, j, k)=E_d$.
\end{itemize}

\begin{figure}[!b]
    \centering
    \begin{subfigure}[t]{0.33\linewidth}
        \centering
        \includegraphics[width=\textwidth]{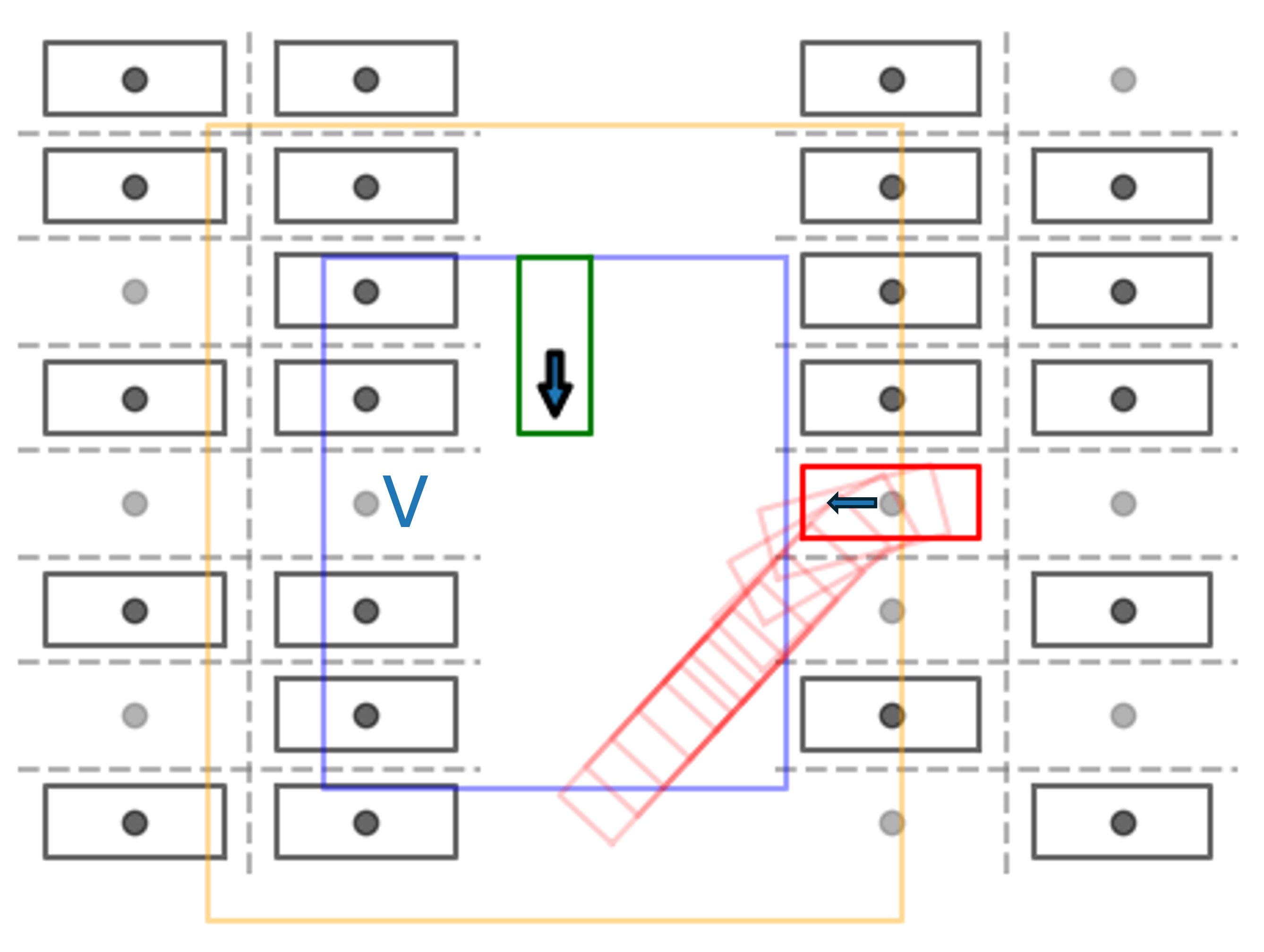}
        \caption{FoV and predictions when red vehicle moves out.}
        \label{fig:exp5_scenario}
    \end{subfigure}
        \begin{subfigure}[t]{0.3\linewidth}
        \centering
\includegraphics[width=\textwidth]{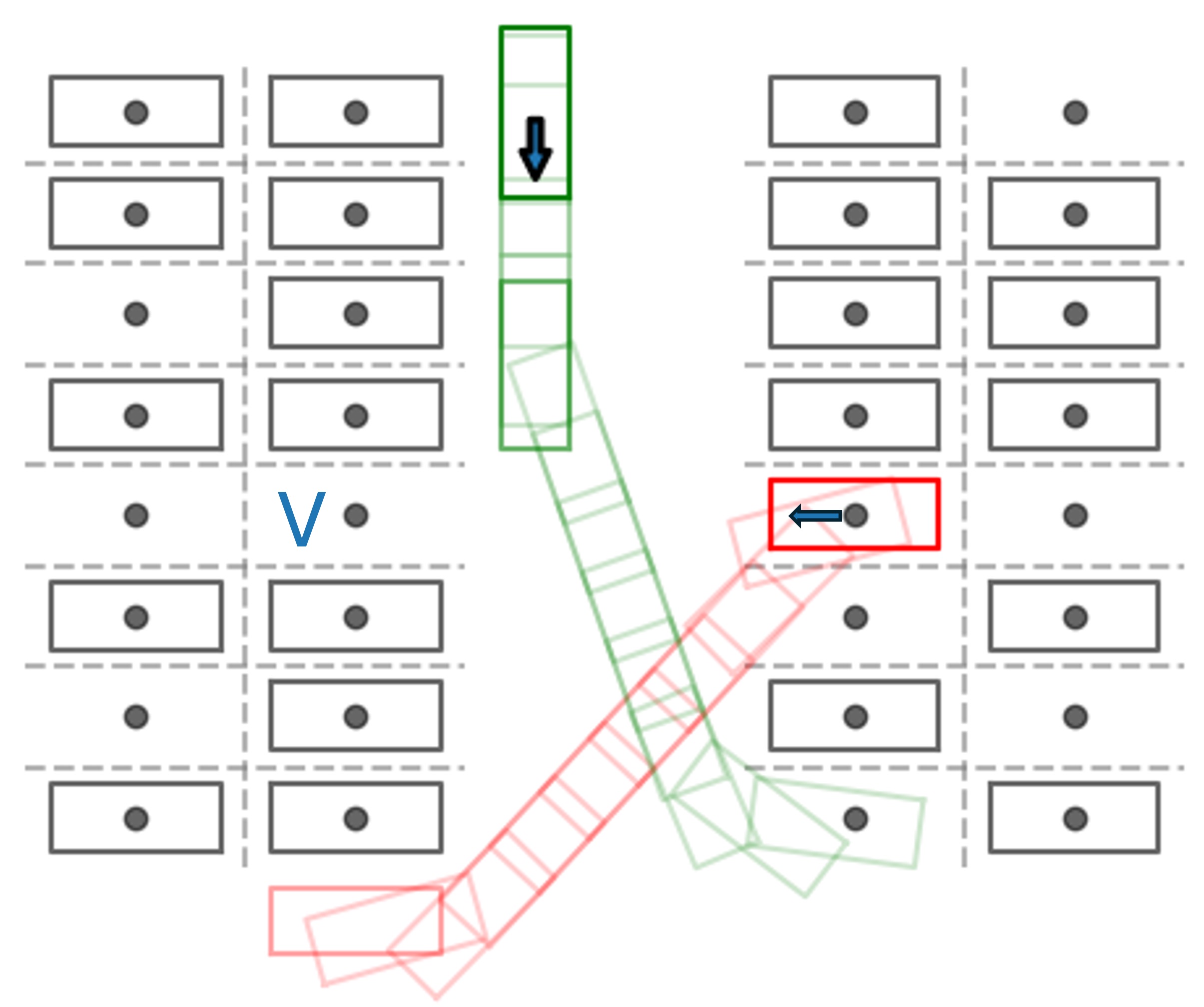}
      \caption{Greedy and Identical prediction.} 
        \label{fig:exp5_traj_same}
    \end{subfigure}   
            \begin{subfigure}[t]{0.3\linewidth}
        \centering
        \includegraphics[width=\textwidth]{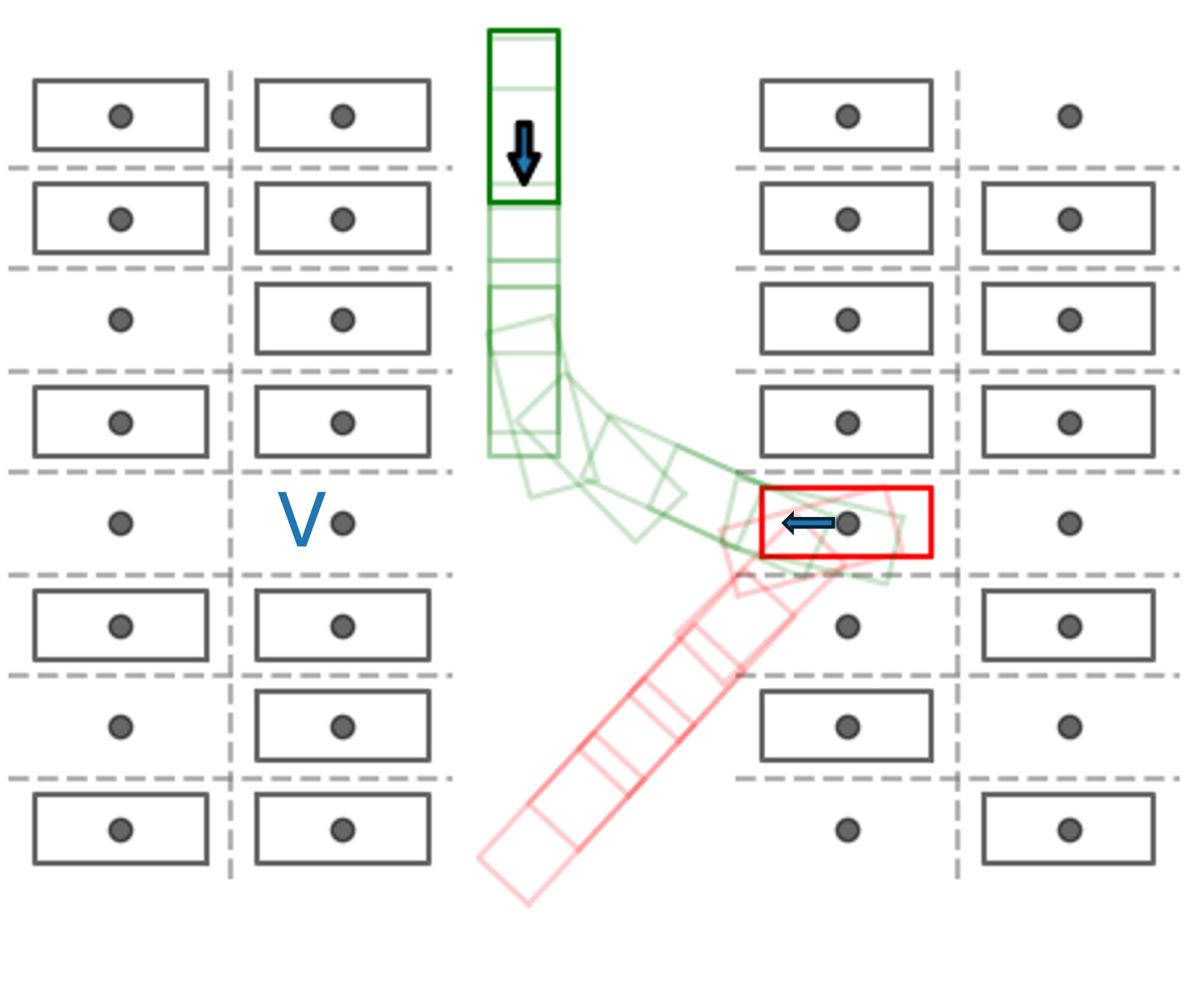}
      \caption{Distinct prediction (Ours).} 
        \label{fig:exp5_traj}
    \end{subfigure}
    
    \begin{subfigure}[t]{0.48\linewidth}
        \centering
        \includegraphics[width=\textwidth]{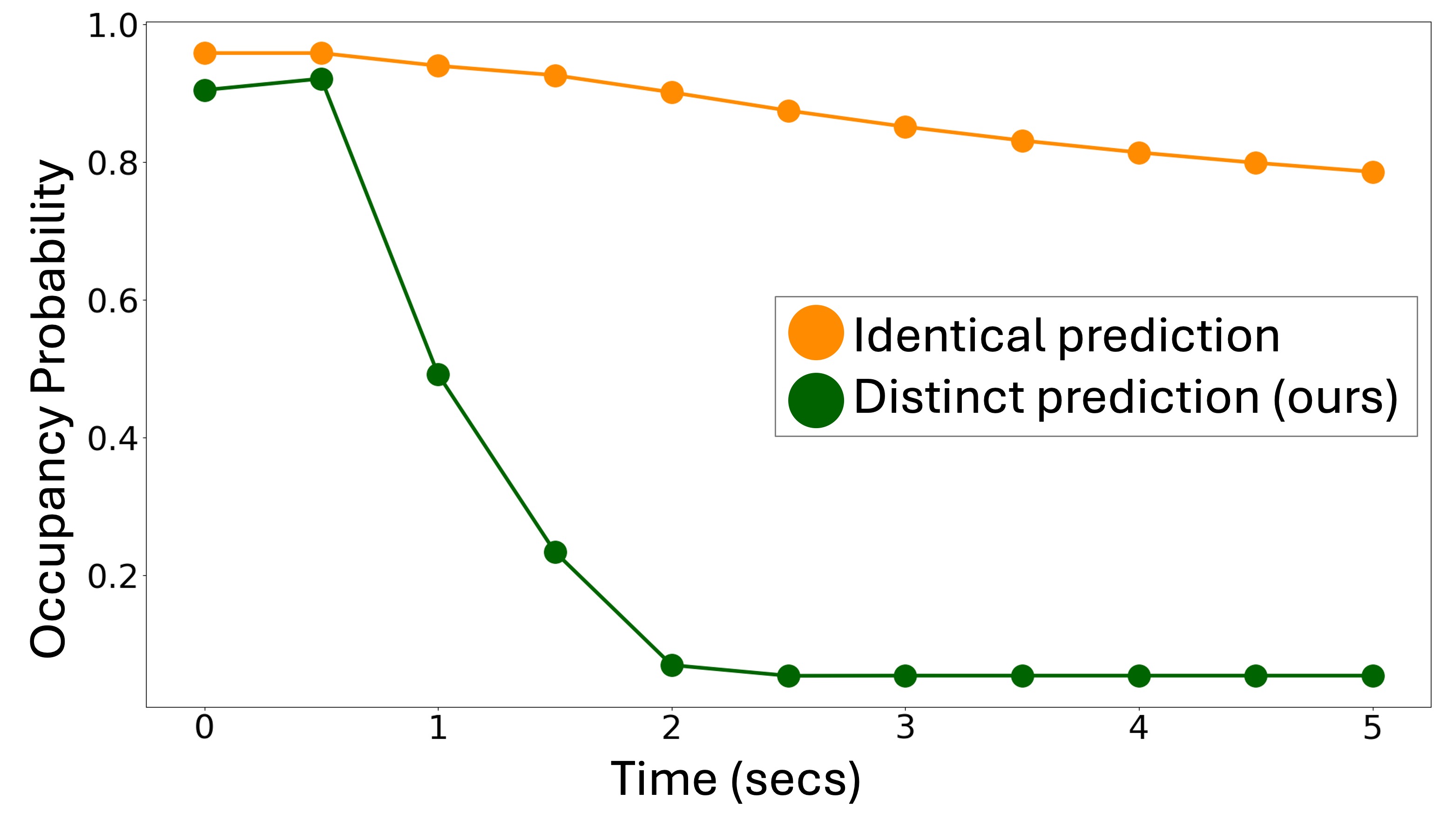}
        \caption{Estimated beliefs of the spot occupied by the red vehicle.}
        \label{fig:exp5_belief_update}
    \end{subfigure} 
        \begin{subfigure}[t]{0.48\linewidth}
        \centering
        \includegraphics[width=\textwidth]{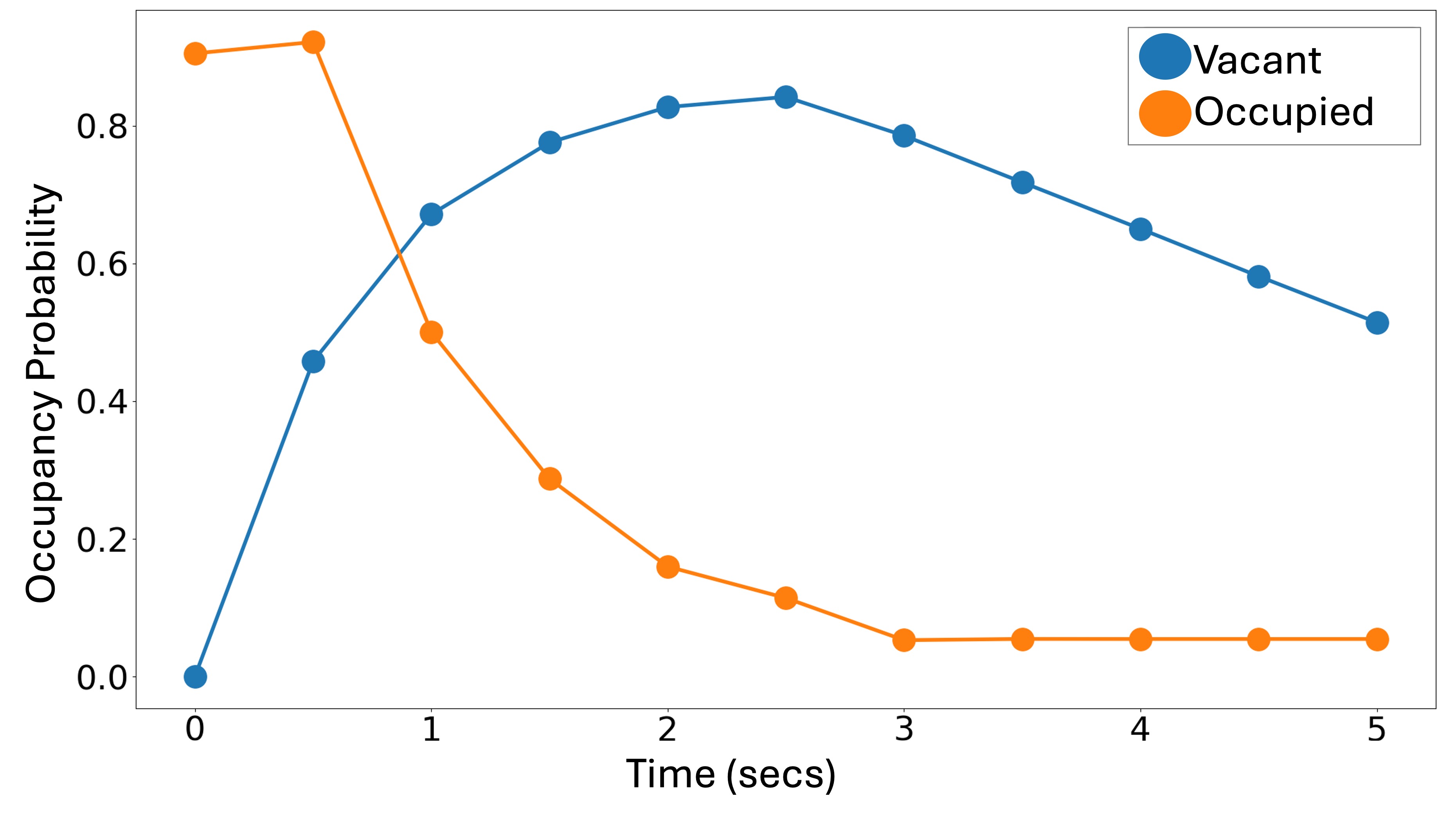}
        \caption{Estimated beliefs of spots that are occupied and vacant.}
        \label{fig:exp5_belief_update_diff}
        \end{subfigure}
    \caption{Comparison of ego vehicle~(green) trajectories~((b) and~(c)) and occupancy probabilities~((d) and (e)) when a vehicle~(red) vacates a spot with a nearby vacant spot~(V).}
    \label{fig:exp5_combined}
\end{figure}

\begin{figure}[!t]
    \centering
    \begin{subfigure}[t]{0.4\linewidth}
        \centering
        \includegraphics[width=\textwidth]{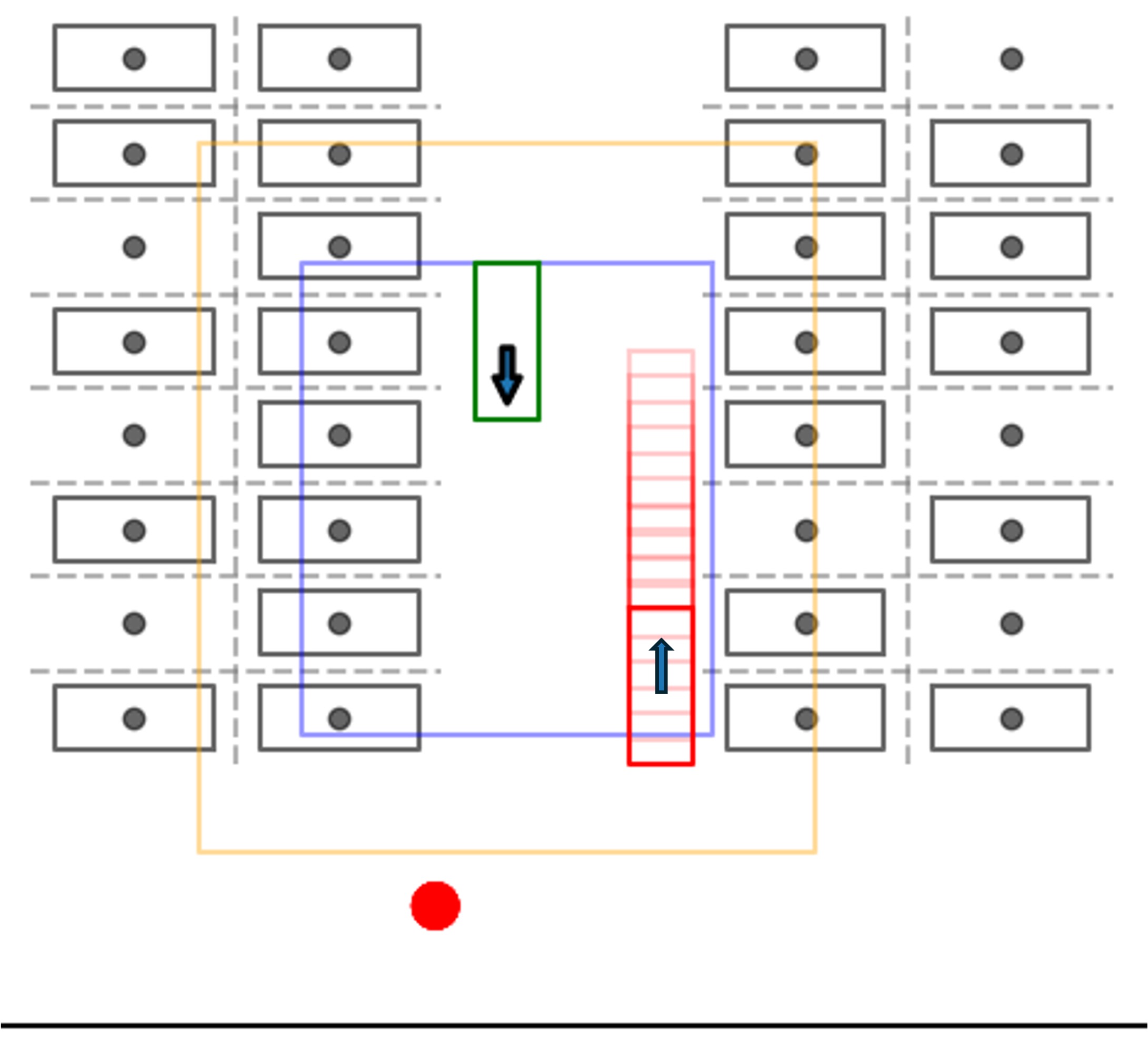}
        \caption{FoV and predictions at initial time.}
        \label{fig:exp6_scenario}
    \end{subfigure}
    \begin{subfigure}[t]{0.57\linewidth}
        \centering
        \includegraphics[width=\textwidth]{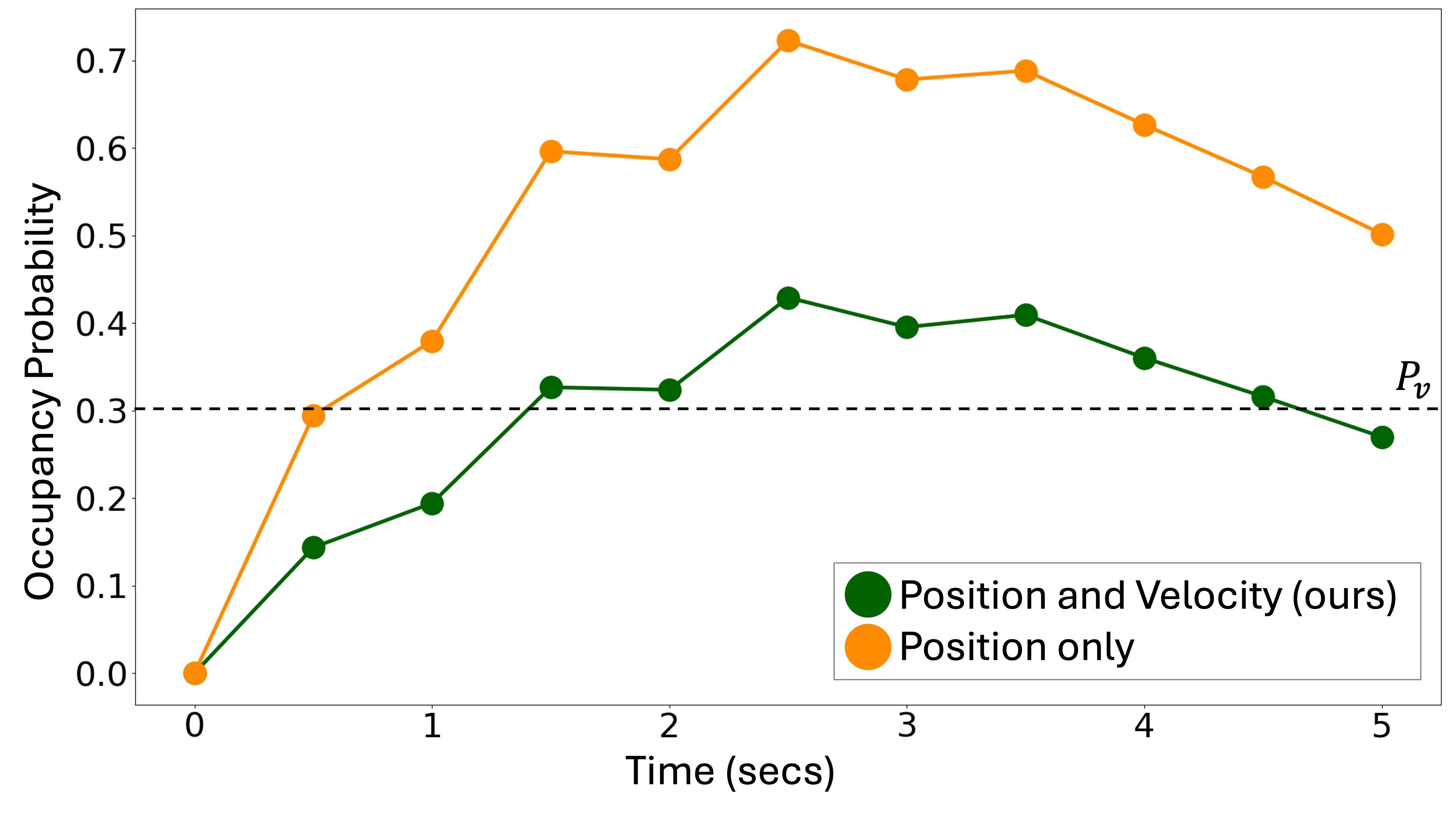}
        \caption{Estimated beliefs of vacant spot.}
        \label{fig:exp6_belief_update}
    \end{subfigure}
    \begin{subfigure}[t]{0.48\linewidth}
        \centering
\includegraphics[width=\textwidth]{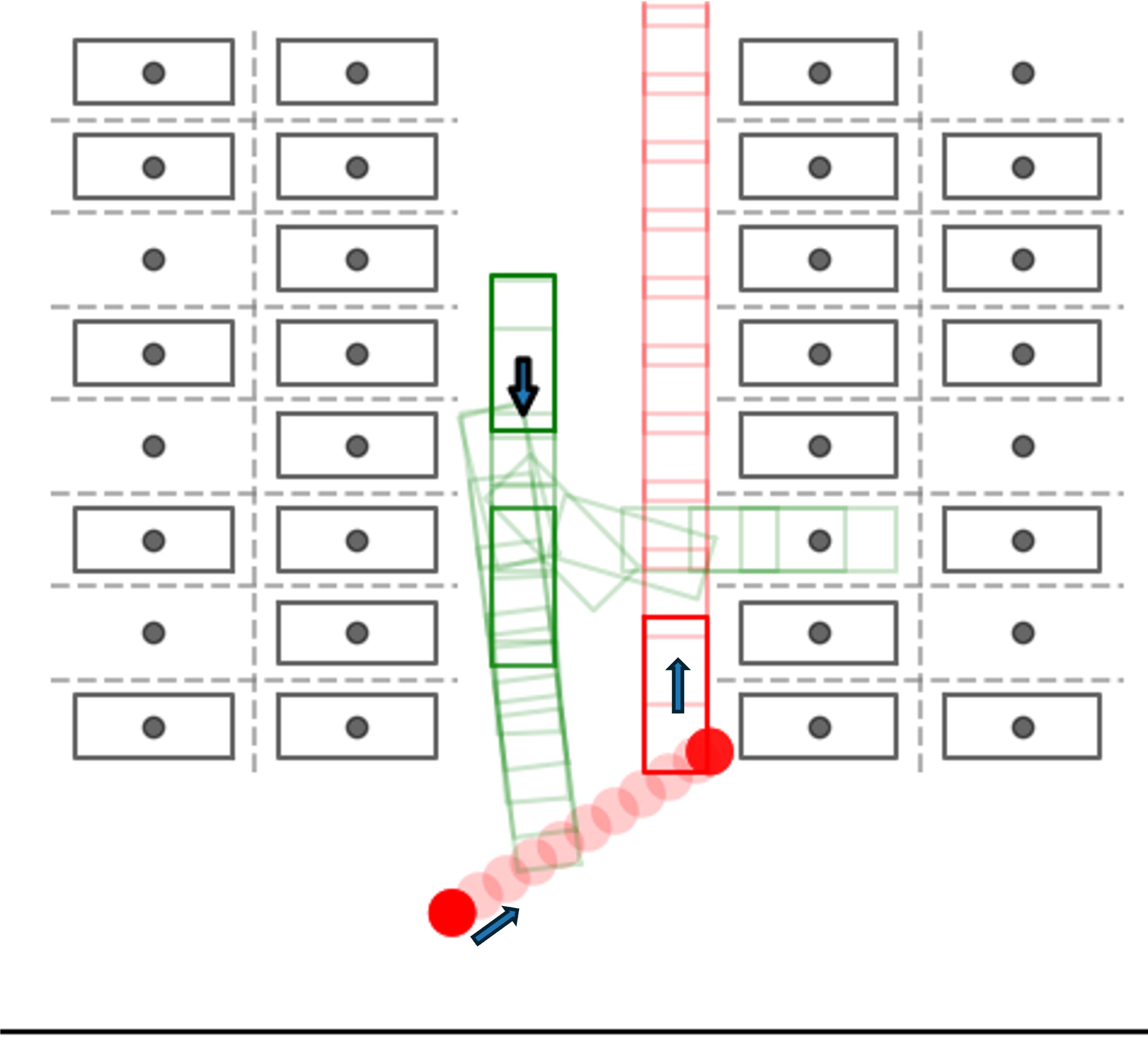}
      \caption{Position only.} 
        \label{fig:exp6_traj_no_vel}
    \end{subfigure}    
            \begin{subfigure}[t]{0.48\linewidth}
        \centering
        \includegraphics[width=\textwidth]{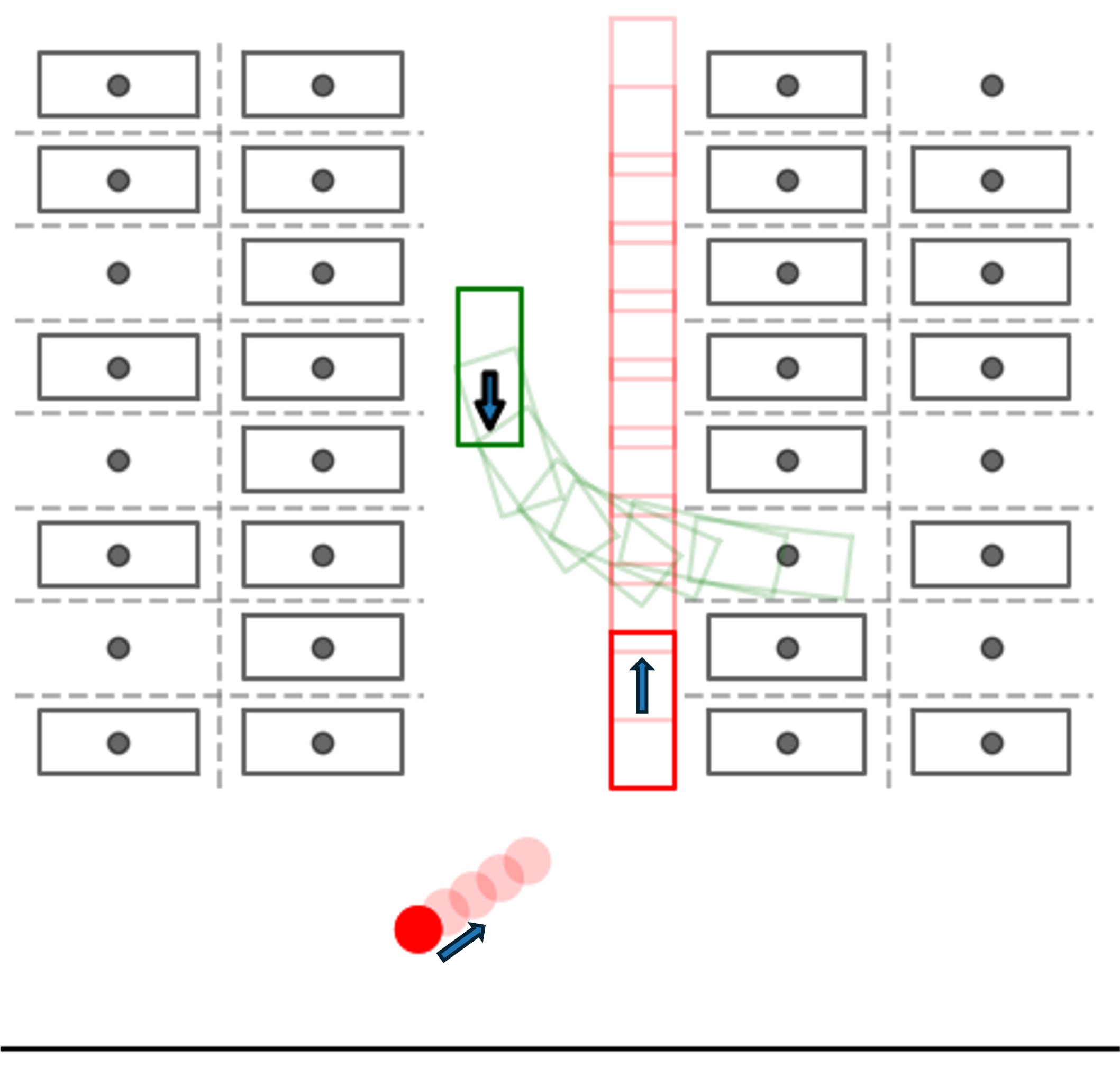}
      \caption{Position and Velocity (Ours).} 
        \label{fig:exp6_traj}
    \end{subfigure}
    \caption{Comparison of ego vehicle~(green) trajectories~((c) and (d)) and occupancy probabilities~(b) when a vehicle~(red rectangle) passes a vacant spot and a pedestrian~(red circle) is entering another vehicle.}
    \label{fig:exp6_combined}
\end{figure}

\begin{figure}[!b]
    \centering
    \vspace{-10pt}
    \begin{subfigure}[t]{0.33\linewidth}
        \centering
        \includegraphics[width=\textwidth]{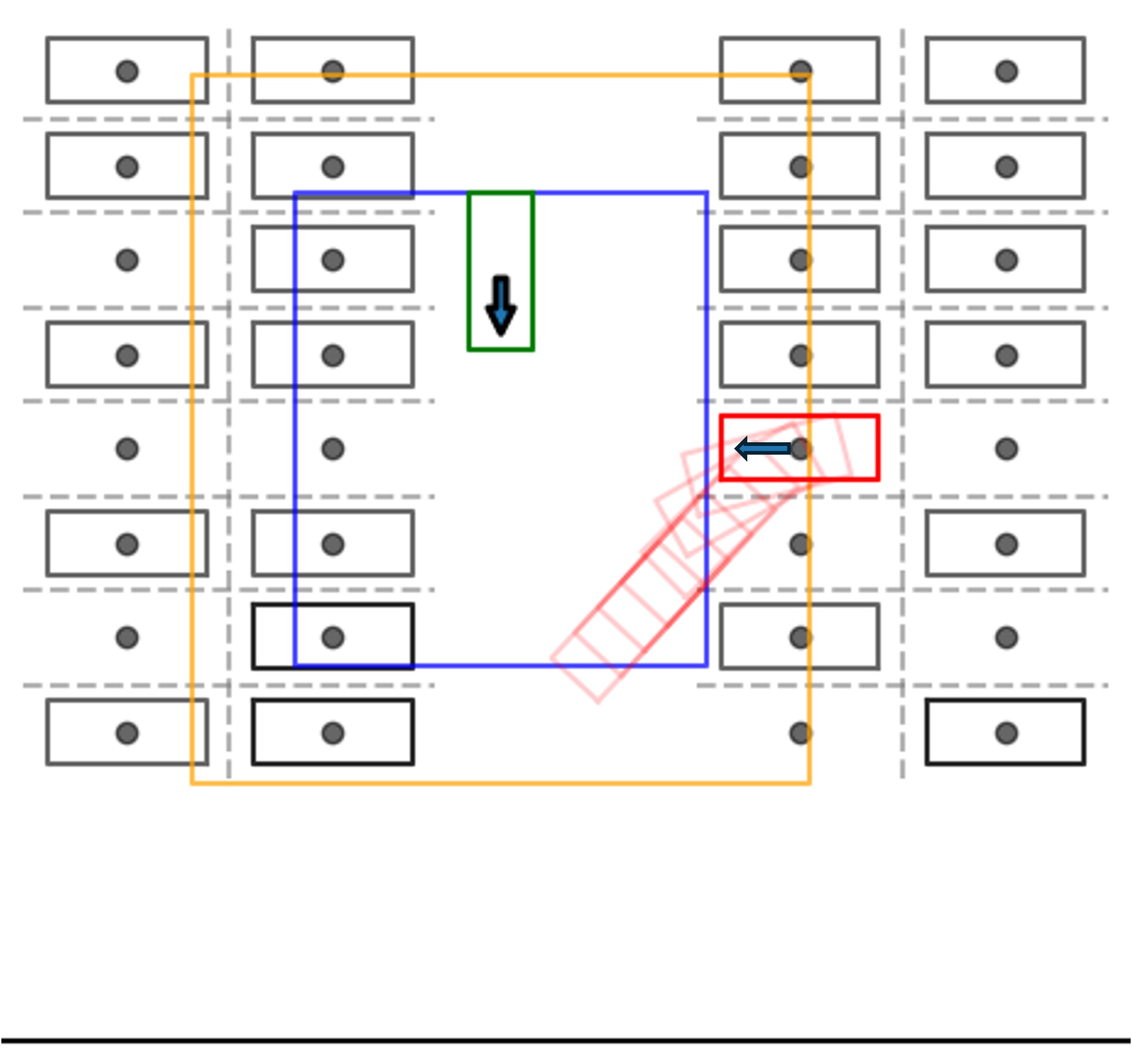}
        \caption{FoV and predictions when red vehicle moves out.}
        \label{fig:wait_scenario}
    \end{subfigure}
    \begin{subfigure}[t]{0.3\linewidth}
        \centering
\includegraphics[width=\textwidth]{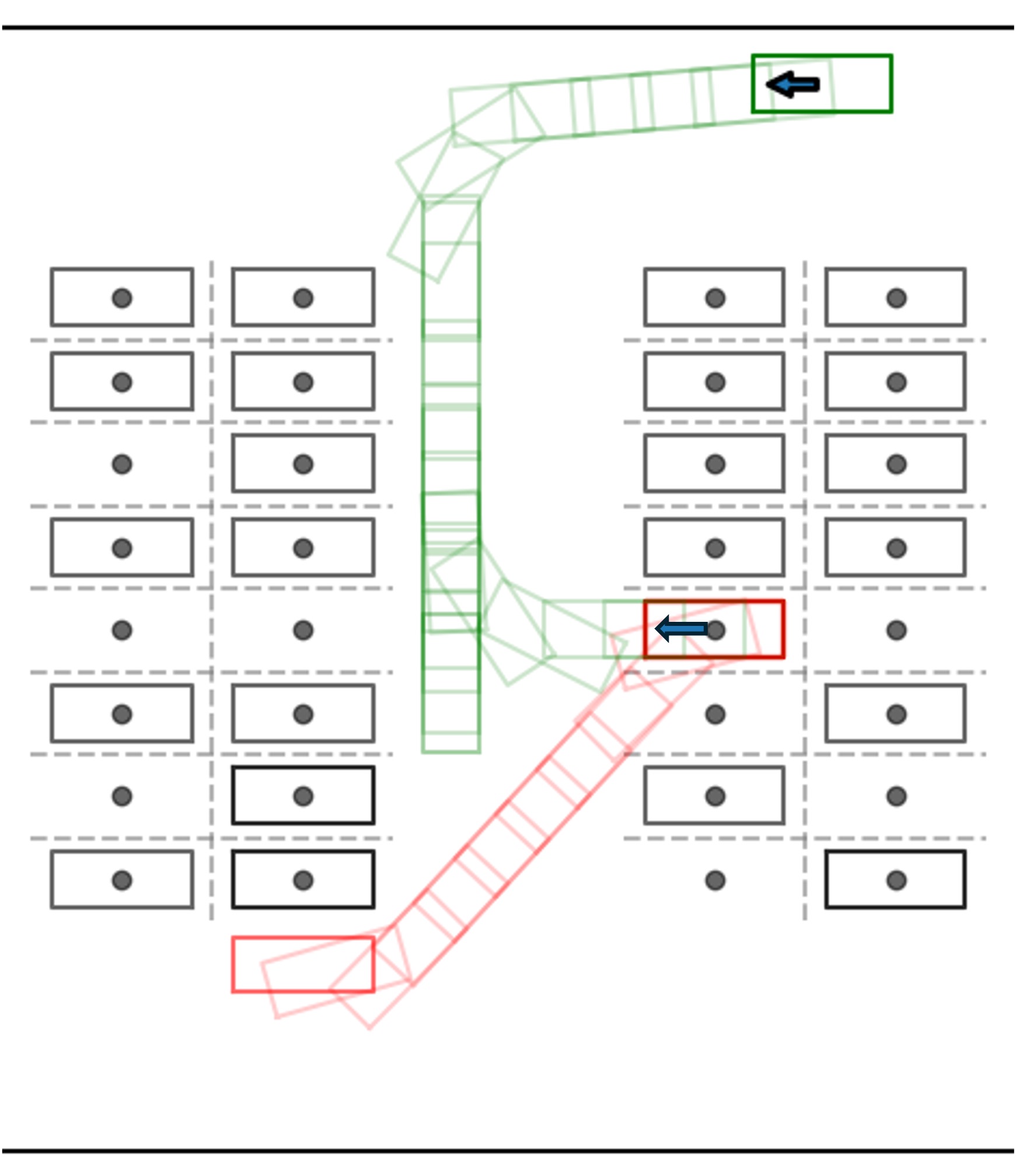}
      \caption{No wait time.} 
        \label{fig:no_wait_traj}
    \end{subfigure} 
                \begin{subfigure}[t]{0.3\linewidth}
        \centering
        \includegraphics[width=\textwidth]{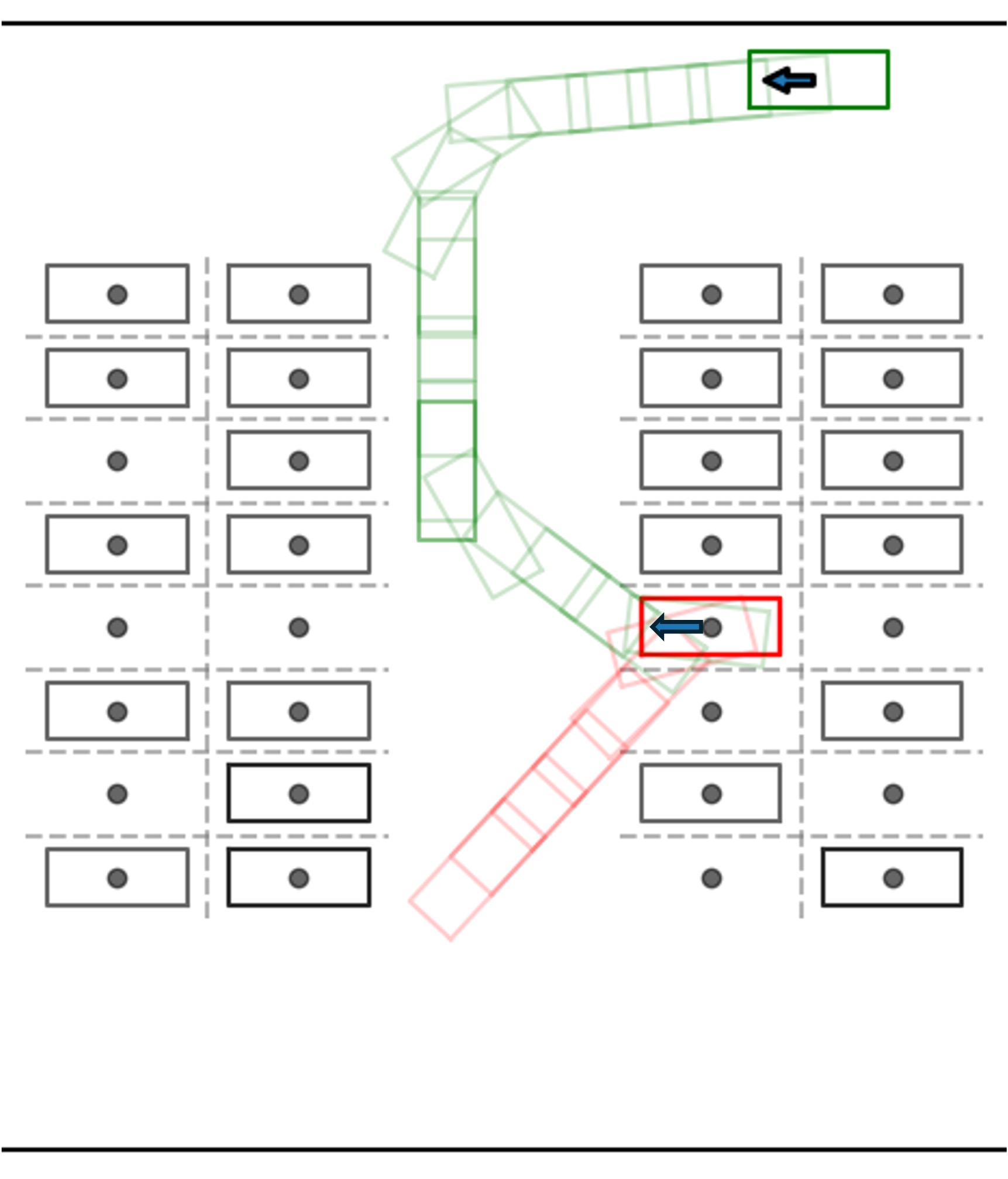}
      \caption{Our strategy.} 
        \label{fig:wait_traj}
    \end{subfigure}
    \caption{Comparison of ego vehicle~(green) trajectories~((b) and (c)) when there is no wait time in~\eqref{wait_go_cost} and another vehicle~(red) moves out of a spot.}
    \label{wait_time}
    \end{figure}

\begin{figure*}[!t]
    \vspace{-15pt}
    \centering
    \begin{subfigure}[t]{0.25\linewidth}
        \centering
        \includegraphics[width=\textwidth]{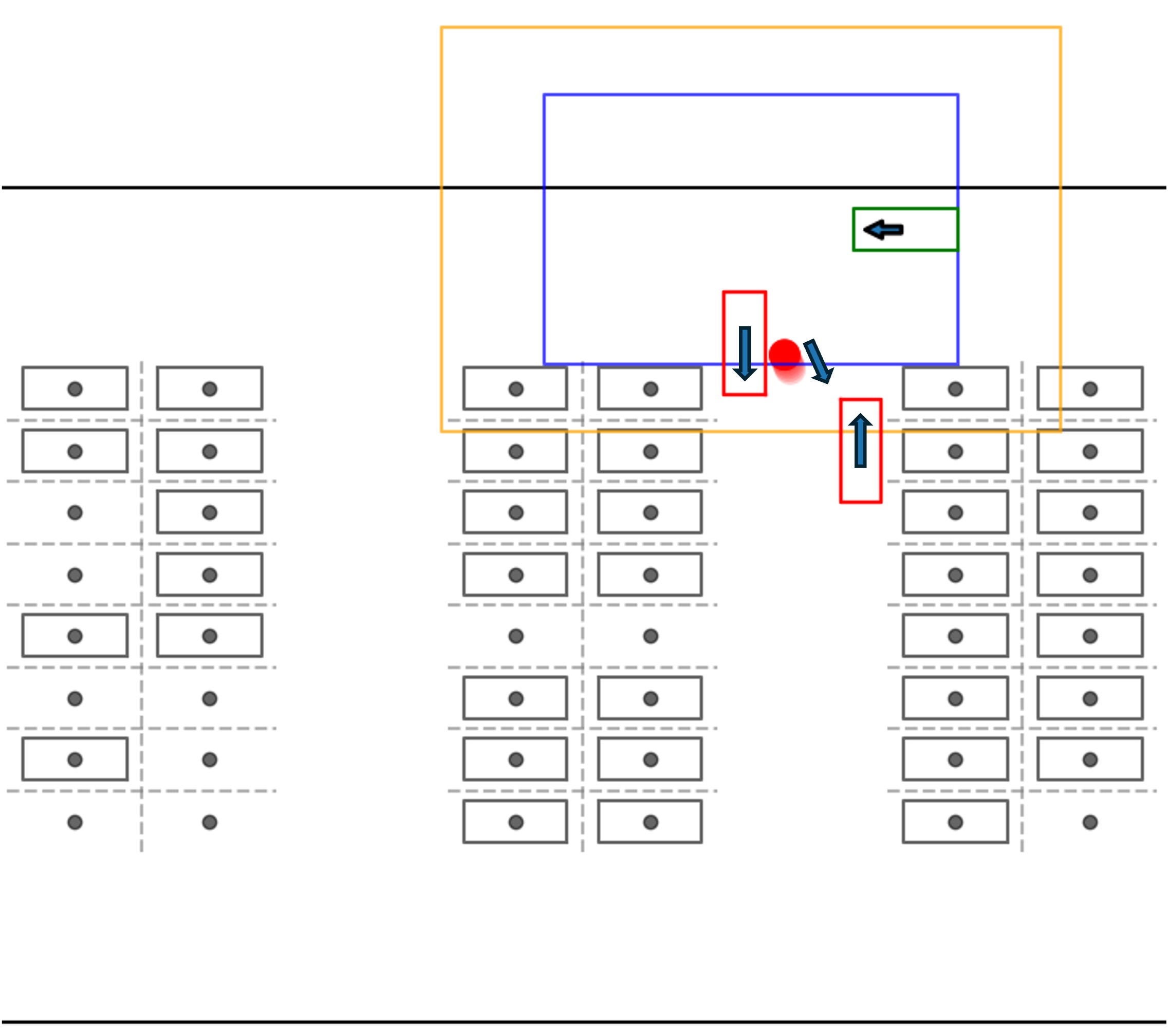}
        \caption{FoV at initial time.}
        \label{fig:exp1_scenario}
    \end{subfigure}
        \begin{subfigure}[t]{0.33\linewidth}
        \centering
\includegraphics[width=\textwidth]{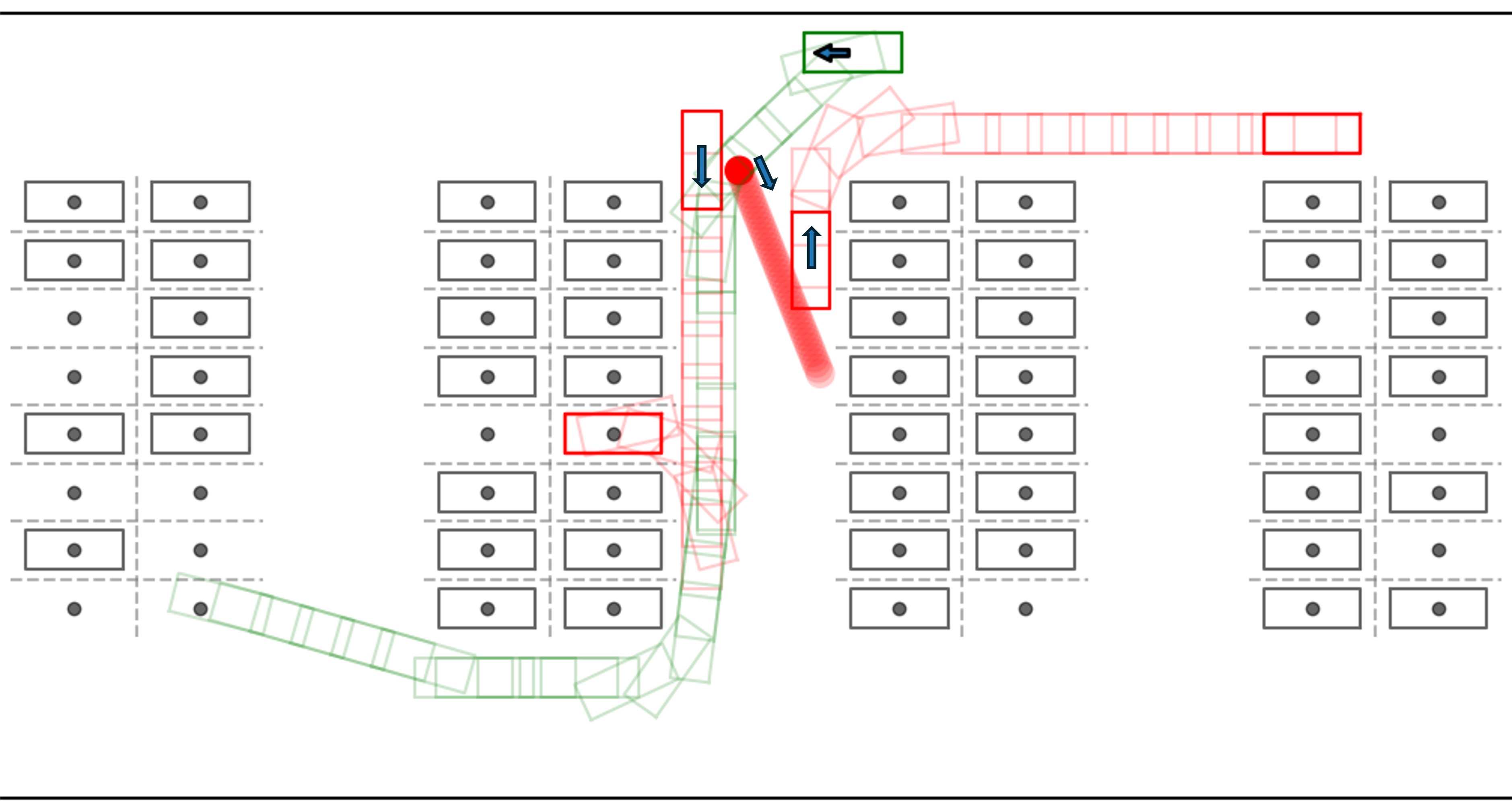}
      \caption{Lawn-mower strategy.} 
        \label{fig:exp1_traj_roomba}
    \end{subfigure} 
        \begin{subfigure}[t]{0.33\linewidth}
        \centering
        \includegraphics[width=\textwidth]{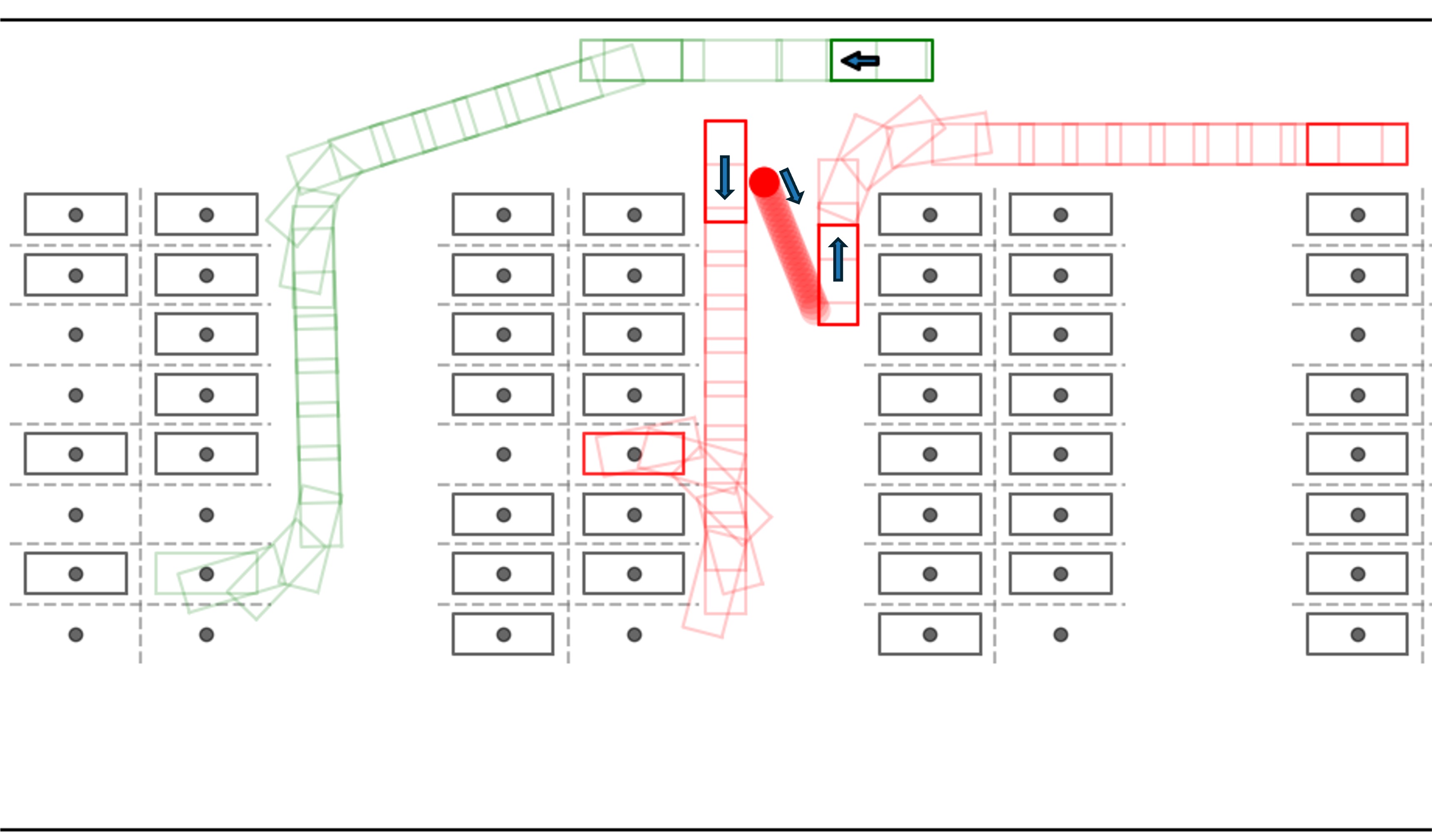}
      \caption{Our strategy planner.} 
        \label{fig:exp1_traj}
    \end{subfigure}   
    \caption{Comparison of trajectories~((b) and (c)) when the ego vehicle~(green) is initially outside the row of parking spots and a row is crowded with other dynamic agents~(red).}
    \label{fig:exp1_combined}
\end{figure*}

Fig.~\ref{fig:exp5_combined} compares our estimator with the Greedy and Identical prediction methods. While Greedy relies only on current occupancy, our approach predicts over a horizon by modeling vacant and occupied dynamics separately, thus lowering occupancy probabilities when a vehicle departs~(Fig.~\ref{fig:exp5_belief_update_diff}) and reducing parking time to 11.5~s versus 16.5~s for others. When the red vehicle passes near the vacant spot V, its occupancy probability temporarily rises but later decreases due to velocity modeling in~\eqref{prob_agent_spot_time}. Although spot V remains vacant, the planner avoids it due to surrounding static vehicles and chooses a more accessible spot.

Fig.~\ref{fig:exp6_combined} compares our estimator with the Position only method, which overestimates the occupancy probability~(Fig.~\ref{fig:exp6_belief_update}) and causes the ego to skip the spot, leading to longer parking times of 24\,s versus 11\,s for our approach. The ego performs a backup maneuver~(Fig.~\ref{fig:exp6_traj_no_vel}) before briefly waiting for a pedestrian initially outside the FoV~(Fig.~\ref{fig:exp6_scenario}).

As shown in Table~\ref{table:metrics}, our complete approach (row 9) outperforms other estimator variants (rows 1--3), reducing parking time by 22\% while maintaining at least 2.3$\times$ greater clearance from dynamic agents.

\subsubsection{Strategy Planner}

We fix the spot occupancy estimator~(Section~\ref{sec:spot_occ_est}) and compare two strategy planner variants against ours~(Section~\ref{sec:strat_planner}):

\begin{itemize}
    \item \textbf{No wait time} strategy always sets $t_w = 0$ in~\eqref{wait_go_cost} so that the ego never pauses near a potentially vacant spot.
        \item \textbf{Lawn-mower} strategy inspired by~\cite{lawn_mower} performs pure exploration~(${\mathcal{C}_e(\tau_e) = - \mathcal{I}(\tau_e)}$ in~\eqref{cost_exp}) when no promising spots are available as classified by~\eqref{filter_future}.
\end{itemize}

Fig.~\ref{wait_time} compares the no wait time strategy with ours. Although our spot occupancy estimator predicts the initially occupied spot to become vacant (Fig.~\ref{fig:exp5_belief_update}), the no wait time strategy first explores further before backing up, resulting in a longer parking time of 27 s versus 18 s for our strategy. Our planner balances waiting and exploration, enabling the ego to park in an initially occupied spot~(Fig.~\ref{fig:wait_traj}) through short, bounded waits, which is effective in dynamic environments.

In Fig.~\ref{fig:exp1_combined}, the Lawn-mower strategy wastes time exploring a crowded row. As a purely exploration-driven strategy, the ego waits for other vehicles, leading to a longer parking time of 43.5 s~(Fig.~\ref{fig:exp1_traj_roomba}). Our strategy planner balances exploration and path efficiency, avoiding detours and reducing parking time to 25.5 s~(Fig.~\ref{fig:exp1_traj}).

Table~\ref{table:metrics} shows that our complete approach (row 9) improves upon other strategy planner variants (rows 4–5), decreasing parking time and path length by at least 39\%, while producing smoother paths, as indicated by lower heading rates and curvature.


\bgroup
\def\arraystretch{1.4}
\centering
\begin{table*}[!t]
\centering
\caption{Planning metrics for 50 randomized experiments across different methods: ablations on \spot{spot occupancy estimator} and \strat{strategy planner}, \baseline{existing work} and \textbf{our approach}. The numbers in \best{green} denote the best value for each metric.}
\begin{tabular}{||p{0.25cm} | p{1.2cm}|p{1.1cm}|p{1.4cm}|p{1.8cm}|p{1.6cm}|p{1.5cm}|p{1.3cm}|p{1.5cm}|p{1.3cm}||}
    \hline \textbf{No.} & 
    \textbf{Method} & 
    \textbf{Runtime} $\downarrow$ (s) & 
    \textbf{Path length} $\downarrow$ (m) & 
    \textbf{Parking time} $\downarrow$ (s) & 
    \textbf{Average distance to closest dynamic obstacle} \newline $\uparrow$ (m) & 
    \textbf{Minimum distance to closest static obstacle} \newline $\uparrow$ (m) & 
    \textbf{Smoothness cost} $\downarrow$ & 
    \textbf{Average heading rate} \newline$\downarrow$ (deg/s) & 
    \textbf{Average curvature} \newline$\downarrow$ (m$^{-1}$) \\
    \hline
    1 & \spot{Greedy} & \best{$0.041 \pm 0.002$} & $34.603 \pm 3.113$ & $23.702 \pm 5.211$ & $3.335 \pm 0.103$ & $0.315 \pm 0.041$ & $4.655 \pm 2.83$ & $12.995 \pm 2.341$ & $0.061 \pm 0.02$ \\
    \hline
    2 & \spot{Identical prediction} & $0.051 \pm 0.002$ & $32.589 \pm 2.773$ & $22.575 \pm 4.097$ & $3.687 \pm 0.074$ & $0.278 \pm 0.092$ & $4.302 \pm 1.12$ & $12.16 \pm 1.154$ & $0.056 \pm 0.002$ \\
    \hline
    3 & \spot{Position only} & $0.049 \pm 0.008$ & $33.735 \pm 5.298$ & $21.539 \pm 1.359$ & $8.331 \pm 0.949$ & $0.314 \pm 0.131$ & $5.045 \pm 2.45$ & $14.6 \pm 2.881$ & $0.058 \pm 0.012$ \\
    \hline
    \hline
    4 & \strat{No wait time} & $0.065 \pm 0.002$ & $45.735 \pm 5.328 $ & $28.524 \pm 2.533$ & $9.228 \pm 1.25$ & $0.264 \pm 0.066$ & $32.668 \pm 4.987$ & $18.989 \pm 3.1$ & $0.058 \pm 0.015$ \\
\hline
    5 & \strat{Lawn-mower} & $0.127 \pm 0.03$ & $41.858 \pm 3.37$ & $32.532 \pm 3.58$ & $3.273 \pm 0.119$ & $0.273 \pm 0.013$ & $12.536 \pm 3.1$ & $17.032 \pm 1.725$ & $0.059 \pm 0.008$ \\
    \hline
    \hline
6 & \baseline{InfPath - Traversal} & $0.059 \pm 0.001$ & $42.364 \pm 3.15$ & $26.725 \pm 5.71$ & $5.232 \pm 0.579$ & $0.261 \pm 0.078$ & $13.583 \pm 1.541$ & $16.86 \pm 2.97$ & $0.061 \pm 0.034$ \\
\hline
7 & \baseline{InfPath - MCBFT} & $0.281 \pm 0.012$ & $35.742 \pm 2.05$ & $23.35 \pm 2.13$ & $6.634 \pm 1.02$ & $0.276 \pm 0.044$ & $14.543 \pm 2.146$ & $14.655 \pm 2.51$ & $0.077 \pm 0.009$ \\
\hline
8 & \baseline{Rule-OBCA} & $0.416 \pm 0.055$ & $32.94 \pm 2.11$ & $25.195 \pm 2.54$ & \best{$9.376 \pm 0.974$} & \best{$0.335 \pm 0.095$} & $20.54 \pm 1.981$ & $15.375 \pm 1.632$ & $0.062 \pm 0.04$ \\
\hline
\hline
    9 & \textbf{Ours} & $0.045 \pm 0.005$ & \best{$24.001 \pm 2.992$} & \best{$17.179 \pm 4.333$} & $8.564 \pm 1.03$ & $0.271 \pm 0.052$ & \best{$3.513 \pm 0.95$} & \best{$11.119 \pm 2.2$} & \best{$0.024 \pm 0.002$} \\
    \hline
\end{tabular}
\label{table:metrics}
\end{table*}
\egroup

\subsection{Comparison with Existing Work}

We benchmark our valet parking framework—combining a spot occupancy estimator and strategy planner—against three existing approaches: \textit{InfPath - Traversal}~\cite{inf_path}, \textit{InfPath - MCBFT}~\cite{inf_path}, and \textit{Rule-OBCA}~\cite{parking_connected}. \textit{InfPath} estimates spot occupancies using a vanilla Bayes filter that ignores dynamic agents and employs an observation model independent of distance from ego vehicle, and presents two planning strategies: \textit{Traversal} (Section~IV-B of~\cite{inf_path}) and Monte Carlo Bayes Filter Tree (MCBFT, Section~IV-C of~\cite{inf_path}). Although~\cite{inf_path} focuses on exploration without dynamic obstacle avoidance, we enforce safety thresholds to discard unsafe paths and select the closest vacant spot according to~\eqref{filter_future}.
\vspace{-2pt}

\begin{itemize}
    \item \textbf{InfPath - Traversal} samples feasible trajectories over a planning horizon of $T = 5 $ s, and selects trajectories that either parks at the closest spot or explores using information gain that is computed using the occupancy estimation method presented in Section~III of~\cite{inf_path}.  
    \item \textbf{InfPath - MCBFT} builds a policy tree from spot occupancy probabilities and selects trajectories using the Upper Confidence Bound. 
    \item \textbf{Rule-OBCA} selects the nearest available spot within the FoV and executes rule-based dynamic collision avoidance using Model Predictive Control with smooth nonlinear constraints~\cite{OBCA}.
\end{itemize}
Quantitatively, rows 6--9 in Table~\ref{table:metrics} show that our approach outperforms existing methods~\cite{inf_path, parking_connected}, reducing path length and parking time by at least 25\% while achieving lower smoothness costs and maintaining safe clearance from obstacle. The runtime presented in Table~\ref{table:metrics} is for our combined spot occupancy estimator and strategy planner.

\vspace{-5pt}
\section{Conclusion and Future Work}
\vspace{-2pt}

We present a trajectory planning framework for autonomous valet parking with limited FoV that integrates a novel \textit{spot occupancy estimator} and a \textit{strategy planner} in uncertain, dynamic environments. We model initially vacant and occupied spots separately to estimate future occupancy probabilities, while also incorporating the motion of dynamic agents. The estimator is integrated with our strategy planner that balances information gain with goal-directed planning, and employs wait-and-go behaviors for promising spots. Simulation results show that our method outperforms existing approaches in efficiency, safety, and smoothness.

The following limitations in our work suggest directions for future research. First, the Hybrid A$^\star$ planner limits steering and velocity samples. Thus, exploring sampling-based methods like MPPI~\cite{mppi} could enable richer control strategies. Second, incorporating intent inference (e.g., parking, yielding, exiting) for other vehicles and evaluating the performance across a wide range of learned trajectory prediction models could enhance interaction-aware planning, thereby improving realism and safety. Third, occlusion-aware sensing models could provide more realistic occupancy estimates. Additionally, developing adaptive waiting behaviors based on lot density may improve efficiency. Finally, implementing our approach in high-fidelity simulators with interactive agents and real-world environments is our future goal.

\vspace{-5pt}


\bibliographystyle{unsrt}
\bibliography{refs}








\end{document}